\pdfoutput=1

\documentclass[11pt]{article}

\usepackage[]{acl}
\usepackage{booktabs}
\usepackage{times}
\usepackage{latexsym}
\usepackage{amsmath} 
\usepackage{longtable}
\usepackage{amsmath}    
\usepackage{amssymb}  
\usepackage{arydshln}
\usepackage{booktabs}

\usepackage[most]{tcolorbox}
\usepackage{xparse}

\usepackage{subcaption}
\usepackage[T1]{fontenc}
\usepackage[ruled,vlined]{algorithm2e}
\usepackage{colortbl}
\usepackage{hyperref} 

\usepackage[utf8]{inputenc}
\usepackage{lipsum}

\usepackage{microtype}
\usepackage{multirow}
\usepackage{graphicx}
\usepackage{amsmath}
\usepackage{inconsolata}

\usepackage{graphicx}
\usepackage{array}
\usepackage{booktabs}
\usepackage{bbding}
\usepackage{pifont}
\usepackage{wasysym}
\usepackage{amssymb}
\usepackage{makecell}
\usepackage[table,dvipsnames]{xcolor} 

\usepackage{amssymb}
\usepackage{pifont}
\usepackage{float}

\newtcolorbox{prompt}[2][]{
    colback=gray!20,
    colframe=white,
    fonttitle=\bfseries\small,
    boxrule=0.4mm,
    fontupper=\small, 
    fontlower=\small,
    coltitle=white,
    title=#2,
    #1,breakable
}
\usepackage{etoolbox}
\usepackage{pgf} 

\definecolor{high}{RGB}{0, 102, 202} 
\definecolor{low}{HTML}{FFFFFF}  

\newcommand{\gradientcell}[6]{%
  \ifdimcomp{#1pt}{>}{#3 pt}{#1}{%
    \ifdimcomp{#1pt}{<}{#2 pt}{#1}{%
      \pgfmathparse{
        int(round(100*((#1-#2)/(#3-#2))))
      }%
      \xdef\tempa{\pgfmathresult}%
      \cellcolor{#5!\tempa!#4!#6} #1%
  }}%
}

\definecolor{highorange}{RGB}{230, 97, 1}   
\definecolor{loworange}{HTML}{FFFFFF}       

\newcommand{\gradientcellorange}[6]{%
  \ifdimcomp{#1pt}{>}{#3 pt}{#1}{%
    \ifdimcomp{#1pt}{<}{#2 pt}{#1}{%
      \pgfmathparse{
        int(round(100*((#1-#2)/(#3-#2))))
      }%
      \xdef\tempa{\pgfmathresult}%
      \cellcolor{#5!\tempa!#4!#6} #1%
  }}%
}

\newlength{\arrayrulewidthOriginal}

\definecolor{warningcolor}{RGB}{250,35,64}
\definecolor{myLightBlue}{RGB}{200,230,255}
\definecolor{myOrange}{RGB}{255, 116, 23}
\definecolor{myBlue}{RGB}{0, 102, 202}
\definecolor{myGreen}{RGB}{0, 153, 102}   
\definecolor{myRed}{RGB}{204, 0, 0}       
\newcommand\graphic{{\includegraphics[width=1.2em]{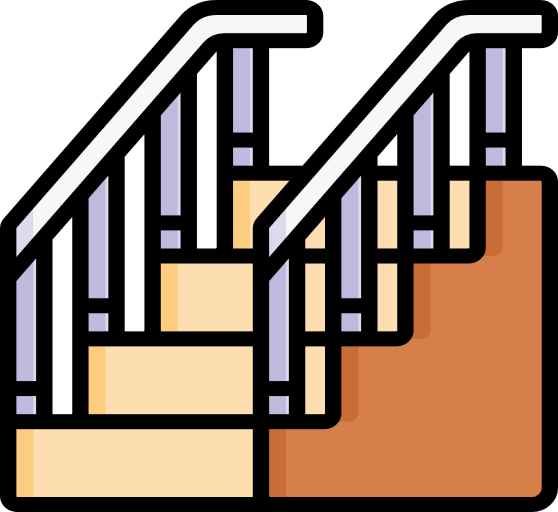}}\space}
\newcommand\STEP{\textsc{Psy-Step}\xspace}
\newcommand\STEPPER{\textsc{Stepper}\xspace}
\newcommand\STEPPERapos{\textsc{Stepper}}
\newcommand\STEPPERSFT{\textsc{Stepper\textsubscript{SFT}}\xspace}
\newcommand\STEPPERDPO{\textsc{Stepper\textsubscript{SFT + Pref.}}\xspace}
\newcommand\STEPPERNoPlan{\textsc{Stepper\textsubscript{SFT\_NoPlan}}\xspace}
\newcommand\GPT{\textsc{GPT-4o}\xspace}
\newcommand\GEMINI{\textsc{gemini-2.0-flash}\xspace}
\newcommand\SMILE{\textsc{SmileChat}\xspace}

\usepackage{colortbl}
\usepackage{xcolor}

\newcommand{\heat}[1]{%
\ifdim #1 pt < 1.6 pt \cellcolor{green!35}#1%
\else\ifdim #1 pt < 1.8 pt \cellcolor{green!20}#1%
\else\ifdim #1 pt < 2.1 pt \cellcolor{yellow!25}#1%
\else\ifdim #1 pt < 2.5 pt \cellcolor{orange!30}#1%
\else \cellcolor{red!30}#1%
\fi\fi\fi\fi
}

\title{\graphic \STEP: Structuring Therapeutic Targets and Action Sequences for Proactive Counseling Dialogue Systems}

\author{
Jihyun Lee\textsuperscript{1},
Yejin Min\textsuperscript{1},
Yejin Jeon \textsuperscript{3,4}\thanks{Work done while at POSTECH.},
SungJun Yang\textsuperscript{2},
\\
\textbf{Hyounghun Kim\textsuperscript{1,2}, 
Gary Geunbae Lee\textsuperscript{1,2}} \\
\\
\textsuperscript{1}Graduate School of Artificial Intelligence, POSTECH \\
\textsuperscript{2}Department of Computer Science and Engineering, POSTECH \\
\textsuperscript{3}MILA  
\textsuperscript{4}McGill University \\
\small\texttt{\{jihyunlee, yeajinmin, sjyang114, h.kim, gblee\}@postech.ac.kr}, \small\texttt{yejin.jeon@mila.quebec}
}

\begin{document}
\maketitle
\begin{abstract}

Cognitive Behavioral Therapy (CBT) aims to identify and restructure automatic negative thoughts pertaining to involuntary interpretations of events, yet existing counseling agents struggle to identify and address them in dialogue settings. To bridge this gap, we introduce \STEP, a dataset that models CBT counseling by explicitly reflecting automatic thoughts alongside dynamic, action-level counseling sequences. Using this dataset, we train \STEPPER, a counseling agent that proactively elicits automatic thoughts and executes cognitively grounded interventions. To further enhance both decision accuracy and empathic responsiveness, we refine \STEPPER through preference learning based on simulated, synthesized counseling sessions. Extensive CBT-aligned evaluations show that \STEPPER delivers more clinically grounded, coherent, and personalized counseling compared to other strong baseline models, and achieves higher counselor competence without inducing emotional disruption.

\end{abstract}


\section{Introduction}

Mental health disorders affect over one billion people worldwide, yet access to treatment remains inadequate. According to global estimates, more than half of individuals with mental disorders do not receive the care they need~\cite{mental_health2}, and the untreated proportion can exceed 75\% in low- and middle-income countries~\cite{mental_health1, who2022mentalhealth}. Contributing factors such as shortages of mental health professionals, limited funding, and persistent social stigma create substantial barriers to care, which motivates growing interest in scalable and complementary approaches, including counseling agents.

\begin{figure}[t]
    \centering
    \includegraphics[width=0.90\columnwidth]{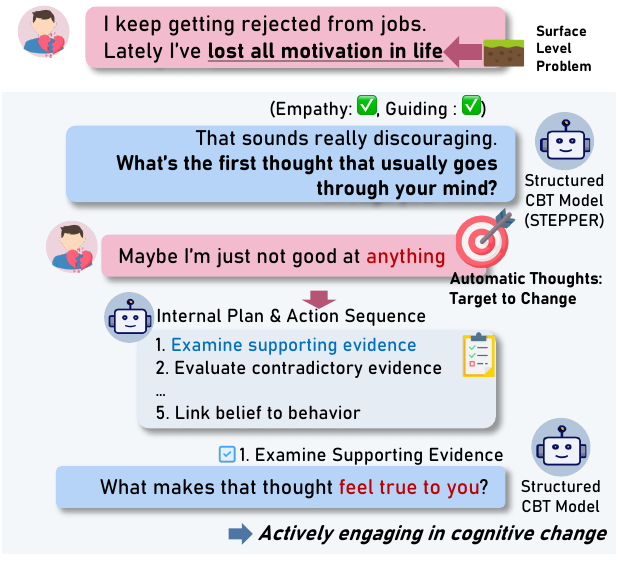}
    \vspace{-5pt}
    \caption{Example of a structured CBT interaction for eliciting automatic thoughts and cognitive reframing.}
    \label{fig:intro}
    \vspace{-10pt}
\end{figure}


Developing effective counseling agents requires high-quality training data, yet collecting real-world counseling conversations is challenging due to privacy concerns and the need for clinical expertise. Recent advances in large language models (LLMs) \cite{gemini, gpt4o, llama3, qwen3} have therefore spurred interest in synthetic dialogue datasets as a scalable alternative. While early synthetic datasets primarily emphasized empathetic responses~\cite{smile}, more recent work incorporate principles from Cognitive Behavioral Therapy (CBT)~\cite{beck2020cbt}, which attributes emotional distress to distorted automatic thoughts that come from the immediate interpretations of events rather than the events themselves. Because these thoughts drive maladaptive emotions and behaviors, identifying and restructuring them is central to achieving meaningful therapeutic change~\cite{healme, cbtLLM, mirror} (Figure~\ref{fig:intro}).


In clinical practice, CBT follows a structured process with two interdependent stages: identification of automatic thoughts that underlie emotional distress, and intervention to modify them~\cite{beck2020cbt, dobson2018cbt}. However, existing CBT-oriented datasets often fail to adequately support this process in two key respects.  First, many prior datasets provide only weak or incomplete specifications of \emph{what} should be treated. Although clients’ negative thoughts are included \cite{maddela-etal-2023-training, cbtLLM}, these datasets frequently conflate surface-level problem descriptions with underlying automatic thoughts. Second, they offer limited guidance on \textit{how} interventions should be carried out. While effective CBT relies on proactive, strategy-specific questioning to elicit and modify automatic thoughts, many datasets lack explicit therapeutic plans or describe only high-level strategies without detailing their execution~\cite{cactus, mirror}. As a result, counseling agents trained on such data tend to produce generic and superficial CBT responses.


In response to these limitations, we introduce \graphic \STEP (\textbf{S}tructured \textbf{T}hought \textbf{E}licitation with \textbf{P}lanning), a dataset designed to support CBT counseling. \STEP makes two key contributions: it explicitly separates \texttt{surface-level} problem expressions from underlying \texttt{automatic thoughts}, enabling accurate identification of the core issues underlying distress, and it defines adaptive therapeutic \texttt{plans} with ordered \texttt{action sequences} to support proactive, strategy-consistent interventions over multi-turn dialogue. Using the proposed \STEP dataset, we train \STEPPER, a counseling agent that proactively elicits automatic thoughts and sequentially executes strategic interventions, and further refine it through preference learning based on simulated client and evaluator feedback.

We comprehensively evaluate \STEPPER across two dimensions: counselor effectiveness and client satisfaction. \STEPPER shows a stronger ability to understand clients’ latent problems by accurately identifying automatic thoughts and producing more guided, strategic counseling behaviors via its explicit plan–action sequence. In terms of client satisfaction, preference alignment is particularly evident: \STEPPER maintains high perceived helpfulness while exhibiting substantially lower hinderness scores, indicating effective intervention without emotional disruption. Expert evaluations further confirm \STEPPERapos’s superior counseling competence and clinical appropriateness.
\section{Related Work}
\vspace{-3pt}

\paragraph{Modeling Client States in Counseling.}
Recent work has sought to better align synthetic counseling datasets with clinical practice by structuring client problem descriptions, drawing on sources such as counseling forums~\cite{smile}, social media~\cite{panic}, and transcribed CBT sessions~\cite{cpsycoun}, and are often augmented with persona-based cognitive distortion labels~\cite{cactus, healme, maddela-etal-2023-training}. However, many approaches still conflate surface-level distress with true therapeutic targets, which results in imprecise representations of core psychological issues. In contrast, \STEP explicitly separates surface-level problems from underlying automatic thoughts, yielding a more faithful representation of CBT’s cognitive targets.\\
\noindent
\textbf{Strategic Control and Action Modeling.}
Prior research on dialogue control spans dialogue skeletons~\citep{soda, chen2025consistentchat}, long-term memory mechanisms~\citep{bae2022keep, jang2024mixed}, and fine-grained action modeling for improved execution reliability~\citep{yao2023react, sun2024pearl}. While some counseling models adopt high-level planning~\citep{cactus, mirror}, they often lack sufficient granularity for clinical execution. \STEP addresses this gap by explicitly encoding ordered action sequences during data generation, enabling precise and strategy-consistent control over dialogue progression.

\begin{figure*}[t]
    \centering
    \includegraphics[width=\textwidth]{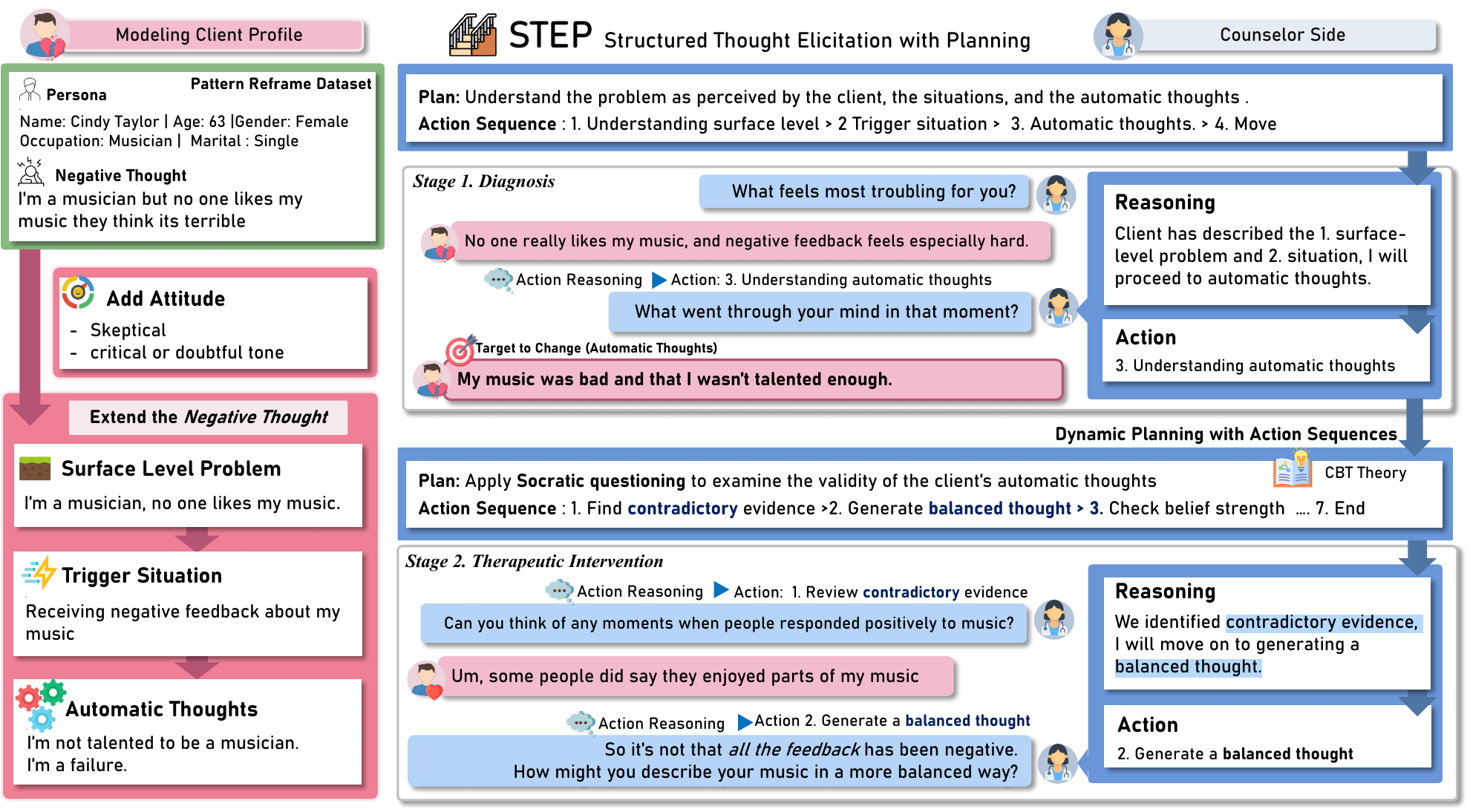}
    \vspace{-20pt}
    \caption{Overview of the \STEP dataset construction and structured CBT counseling flow.
The figure illustrates how client profiles are modeled, how surface-level problems and automatic thoughts are elicited during the diagnostic stage, and how structured action sequences guide therapeutic interventions through stepwise CBT reasoning.}
    \label{fig:main}
    \vspace{-13pt}
\end{figure*}

\vspace{-5pt}
\section{\graphic\textsc \STEP: Structured Counseling Dataset for CBT}
\vspace{-3pt}
In this section, we introduce \STEP, a structured counseling dataset designed to support CBT by explicitly modeling and addressing maladaptive automatic thoughts. We posit that effective counseling data should (1) capture both surface-level distress and the underlying automatic thoughts, and (2) enable proactive, plan-guided counseling to effectively modify such thoughts. To this end, \STEP is constructed with three key design principles: (i) client profiles that characterize surface-level problems and underlying automatic thoughts, (ii) clear separation between diagnostic interviews and therapeutic stages to reflect their distinct roles, and (iii) proactive counseling guided by predefined stage-specific therapeutic plans and action sequences. The overall construction process is illustrated in Figure~\ref{fig:main}, and the specific prompts and implementation details are provided in Appendices~\ref{app:step_detail} and ~\ref{app:prompt_for_step}. \textsc{GPT-4o-mini} \cite{gpt4omini} is primarily used for dialogue synthesis.

\subsection{Client Profile Construction}
\label{sec:step_profile}
We first construct client profiles that capture clinically relevant information. To this end, the \textsc{PatternReframe} dataset \cite{maddela-etal-2023-training} is utilized as the primary source, in which human annotators assign \texttt{negative thoughts} to individual personas. While these annotations provide realistic negative thoughts, they do not explicitly distinguish automatic thoughts from surface-level problem descriptions, and the two are often conflated or insufficiently specified. To address this limitation, we apply targeted prompting to decompose each negative thought into two distinct components: \texttt{surface-level problem} that represent observable and consciously reported distress, and \texttt{automatic thought}, which is defined as the unconscious, involuntary interpretation. To further enrich the conversational context, we generate a situational description which specifies the \texttt{triggering circumstances} based on persona, along with the client’s \texttt{attitude} towards counseling. An example of the resulting profile is provided in Appendix~\ref{app:step_profile}.



\subsection{Planning and Dialogue Generation}

Using the constructed client profiles, we generate multi-turn counseling dialogues. Each dialogue consists of two sequential stages—a \texttt{diagnostic stage} and an \texttt{intervention stage}—with each stage governed by a stage-specific therapeutic plan and an ordered action sequence. The diagnostic stage focuses on eliciting the client’s latent automatic thoughts through guided questioning, while the intervention stage focuses on reframing these thoughts by applying the corresponding therapeutic plan and action sequence. We adopt a script-based generation paradigm inspired by \citet{cactus}, in which the dialogue for each stage is generated within a single prompt to ensure global coherence and adherence to the intended plan.

\paragraph{Plan and Action Sequence Construction.}

Before dialogue generation, a therapeutic plan and an ordered action sequence are defined for each stage.
Here, a \emph{therapeutic plan} specifies the high-level counseling objective and strategy for a stage, while the \emph{action sequence} operationalizes the plan as an ordered set of concrete, observable counselor actions to be executed during dialogue (see Figure~\ref{fig:main} for an example). In the \texttt{diagnostic stage}, the counselor does not yet know what problems the client brings to the session; therefore, we employ a predefined plan and action sequence designed to systematically elicit the client’s presenting problems and underlying automatic thoughts through guided questioning. In contrast, after the \texttt{diagnostic stage}, we adopt a dynamic planning strategy in the \texttt{intervention stage}. Here, the LLM generates a therapeutic plan and an ordered action sequence based on the identified presenting problem and automatic thoughts from previous stage. Specifically, the model is conditioned on the surface-level problem, situational trigger context, automatic thoughts, and a predefined set of CBT strategies, producing a client-specific plan with 5–7 concrete action steps. CBT strategies and example therapeutic plans with action sequences are provided in Appendix~\ref{app:CBT_strategy} and ~\ref{app:step_plan_example}.

\paragraph{Dialogue Generation.} In both stages, each dialogue is generated as a sequence of alternating counselor and client turns,
$D = \{(t_1^{\text{coun}}, t_1^{\text{cli}}), \ldots, (t_T^{\text{coun}}, t_T^{\text{cli}})\}$ conditioned on the client profile, therapeutic plan, and action sequence. For the \texttt{intervention} stage, the dialogue is additionally conditioned on the history from the preceding \texttt{diagnostic} stage. Each counselor turn $t_i^{\text{coun}}$ comprises three components: an internal \texttt{action reasoning} step
$r_i^{\text{act}}$ that determines the appropriate counseling action, an \texttt{action indicator} $a_i$ reflecting
progress within the stage-specific plan, and a natural-language \texttt{utterance} $u_i^{\text{coun}}$ delivered to the client.
To prevent action skipping, the LLM is explicitly prompted to follow a predefined action sequence, and only advances to the next step when the current objective is met or repeats the current action when additional probing is required. Client turns $t_i^{\text{cli}}$ are generated analogously, and consists of \texttt{internal reasoning}
$r_i^{\text{cli}}$ and  \texttt{utterance} $u_i^{\text{cli}}$.


\subsection{Dialogue Filtering and Quality Control}
To ensure therapeutic validity, we filter dialogues based on CBT fidelity and plan adherence. Dialogues are retained only if they (i) achieve acceptable quality under the Cognitive Therapy Rating Scale (CTRS) \cite{ctrs}, which evaluates therapeutic skills, and (ii) follow the prescribed intervention plan without skipping or misordering actions. CBT fidelity is assessed using \GPT, and dialogues with any CTRS item scored at 4 or below (on a 6-point scale) are discarded. After filtering, 67.71\% of dialogues are retained, which yields 6,425 dialogues with 231,172 turns.

\subsection{Expert Review of Dataset}
Three mental health professionals conducted evaluations for 130 randomly sampled dialogues. Each dialogue was rated on a 5-point scale (1 = very poor, 5 = excellent) across six dimensions: coherence between surface-level problems and automatic thoughts, surface problem coverage, automatic thought elicitation, plan–action appropriateness, action execution fidelity, and interpersonal effectiveness. The average scores were 4.92, 4.90, 4.96, 4.83, 4.89, and 4.88, respectively, which indicates consistently high quality across all criteria. Further details of the human evaluation protocol are provided in Appendix~\ref{app:human_dataset_quality}.

\subsection{Comparison \graphic\STEP with Existing Counseling Datasets}
\begin{table*}[t!]
\centering
\resizebox{\textwidth}{!}{
\begin{tabular}{@{}lccccccccc@{}}
\toprule
\textbf{Dataset} 
& \textbf{Counseling Theory} 
& \textbf{Problem Representation} 
& \textbf{Problem Source} 
& \textbf{Intervention Structure} 
& \textbf{Open} 
& \textbf{Language} 
& \textbf{\# of Dialogues} 
& \textbf{Avg. Turns} \\
\midrule

\textsc{PsyCon}~\cite{psycon} 
& Not Specified 
& Disorder-specific Experiences 
& Online Forum 
& None 
& $\triangle$ 
& English 
& 1,020 
& 24.6 \\

\SMILE~\cite{smile} 
& Not Specified 
& Mental Health Questions 
& Online Q\&A Platforms 
& None 
& Yes 
& Chinese 
& 55,165 
& 10.4 \\

\textsc{Psych8k}~\cite{psych8k} 
& Cognitive Behavioral Therapy  + Others
& Patient-reported Concerns
& Counseling Records 
& None 
& Yes 
& English 
& 8,187 
& 10.0 \\

HealMe~\cite{healme} 
& Cognitive Behavioral Therapy 
& Negative Thoughts 
& Crowdsourced 
& Planning 
& No 
& English 
& 1,300 
& 3.0 \\

CBT-LLM~\cite{cbtLLM} 
& Cognitive Behavioral Therapy 
& Mental Health Questions 
& Online Q\&A Platforms 
& Planning 
& No 
& Chinese 
& 22,327 
& 1.0 \\

CACTUS~\cite{cactus} 
& Cognitive Behavioral Therapy 
& Negative Thoughts 
& Crowdsourced 
& Planning 
& Yes 
& English 
& 31,577 
& 16.6 \\
\cdashline{1-9}

\graphic \STEP
& Cognitive Behavioral Therapy 
& \textbf{\makecell[c]{Surface-level \\ + Automatic Thoughts}}
& Crowdsourced 
& \textbf{\makecell[c]{Planning \\ + Action Sequence}}
& Yes 
& English 
& 6,425 
& \textbf{18.0} \\

\bottomrule
\end{tabular}
}
\vspace{-5pt}
\caption{Comparison of counseling dialogue datasets across theory, structure, and problem representation.}
\label{tab:dataset_comparison}
\vspace{-10pt}
\end{table*}

Table~\ref{tab:dataset_comparison} compares \STEP with existing counseling dialogue datasets.
While prior datasets primarily focus on client-reported problems, \STEP explicitly models both surface-level problems and underlying automatic thoughts, enabling deeper cognitive exploration.
Moreover, unlike previous datasets that provide high-level planning, \STEP incorporates explicit action sequences, which supports coherent and robust counseling over extended multi-turn interactions.
Accordingly, \STEP exhibits substantially longer dialogues compared to other datasets, which reflects its step-wise intervention structure.

\section{\textsc{Stepper}: Structured  CBT Counseling Model}
\vspace{-5pt}
\label{sec:stepper}

\begin{figure}[t]
    \centering
    \includegraphics[width=0.93\columnwidth]{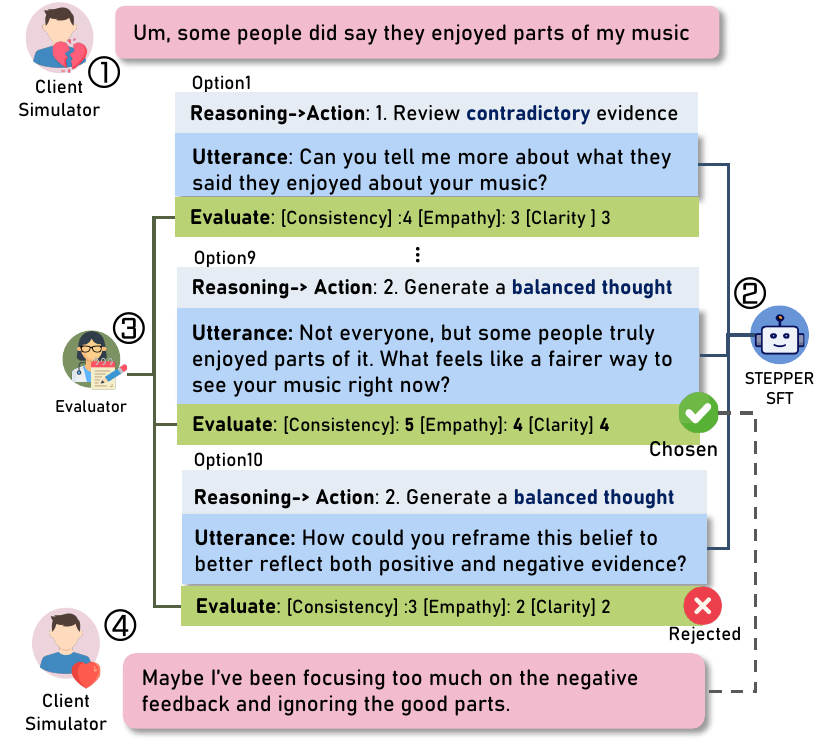}
    \vspace{-5pt}
    \caption{Illustration of the simulation-based process for collecting preference data for utterance selection.}
    \label{fig:dpo}
    \vspace{-15pt}
\end{figure}

\textbf{Supervised Fine-Tuning.}
\label{sec:simulation}
Using the generated \STEP dataset, we train our structured CBT counseling model \textsc{Stepper} via supervised fine-tuning with parameter-efficient Low-Rank Adaptation (LoRA)~\cite{lora}. The model employs two task-specific adapters: an \emph{utterance adapter} and a \emph{planner adapter}. For the utterance adapter, at each turn $t$, the model conditions on the previous dialogue context $D_{t-1}$, the previous counseling action $act_{t-1}$, and the next action candidate following the current action sequence. It generates (1) an internal reasoning trace $r^{\text{act}}_t$ to determine whether to transition to the next action or reiterate the current one, (2) the finalized action decision $act_t$, and (3) the counselor’s actual response $u^{coun}_t$ that corresponds to that finalized action. On the other hand, the planner adapter performs planning and action sequence generation. Given a diagnostic dialogue $D^{\text{diag}}$, the model is trained to generate a therapeutic plan along with an ordered sequence of counseling actions grounded in the diagnosis.

\noindent
\paragraph{Preference Tuning.}
To enhance the counseling ability of \textsc{Stepper}, we further refine the model via preference learning, focusing on empathy and plan adherence for the utterance adapter, and plan completeness and feasibility for the planner adapter.
Preference signals are collected in a counseling simulation that includes an \textsc{Client} simulator conditioned on a client profile (\S~\ref{sec:step_profile}),  the \textsc{Stepper} model initialized with supervised fine-tuning, and an \textsc{Evaluator} that scores model outputs using given metrics. Each simulation proceeds in four steps (Figure~\ref{fig:dpo}): (1) the client simulator produces a user utterance based on client profile and dialogue history; (2) \textsc{Stepper} generates $N=10$ candidate responses via stochastic search; (3) the evaluator scores all candidates based on the metrics; and (4) the highest-scoring response is selected as the final output, while the top two and the worst two candidates are paired to construct preferred and rejected samples for Direct Preference Optimization (DPO). 

For utterance-level alignment, candidates are scored on \emph{action consistency}, \emph{empathy}, and \emph{clarity}; for planner alignment, candidate plans are evaluated based on \emph{completeness}, \emph{feasibility}, and \emph{plan--action alignment}. In all cases, scores are assigned on a 1--5 scale and averaged to determine preference rankings. From this simulation, we obtain 26,576 preference pairs for the utterance adapter and 6,136 for the planner adapter. In this process, \GPT is used to instantiate both the client simulator and the evaluator. We provide collected preference examples, human validation for the selected and rejected candidates, and the used prompts in Appendices~\ref{app:stepper_detail} and ~\ref{app:prompt_for_stepper}.


\begin{table*}[t]
\centering
\resizebox{\textwidth}{!}{%
\begin{tabular}{lcccccccc}
\toprule 
\multirow{2}{*}{\textbf{Model}} & 
\multicolumn{3}{c}{\textbf{General Skills ($\uparrow$)}} & 
\multicolumn{4}{c}{\textbf{CBT-specific Skills ($\uparrow$)}} & 
\multirow{2}{*}{\makecell{Ques.-Ref.  \\ \textbf{Diversity} ($\uparrow$)}} \\ 

\cmidrule(lr){2-4} \cmidrule(lr){5-8} 

& \textbf{Understand} 
& \textbf{Interpers.} 
& \textbf{Collabo.} 
& \textbf{Guided Dis.} 
& \textbf{Focus} 
& \textbf{Strategy} 
& \textbf{AT. Coverage} 
\\
\midrule
\textsc{GPT-4o} 
& 3.74 & 5.63 & 5.04 & 3.94 & 3.73 & 2.58 & 2.49 & 1.07 \\

\GEMINI 
& 3.78 & 5.23 & 4.35 & 3.73 & 3.81 & 3.35 & 2.82 & 1.42 \\

\textsc{SmileChat} 
& 2.62 & 4.18 & 3.22 & 2.71 & 3.15 & 2.45 & 1.54 & 1.31 \\

\textsc{CBT-LLM} 
& 2.93 & 4.13 & 2.37 & 2.66 & 3.71 & 3.29 & 2.95 & 1.66 \\

\textsc{Llama-Psych8k} 
& 4.01 & 5.74 & 5.62 & 4.58 & 4.66 & 4.59 & 3.77 & 1.62 \\

\textsc{Camel} 
& 4.51 & 5.75 & 5.48 & 4.63 & 4.73 & 4.49 & 4.69 & 1.88 \\

\cdashline{1-9}

\STEPPERNoPlan
& 3.96 & 5.64 & 5.04 & 4.14 & 4.14 & 3.15 & 4.08 & 1.93 \\

\rowcolor{gray!15}\STEPPERSFT
& 4.70 & 5.81 & 5.62 & \textbf{5.00} & 5.01 & 4.69 & \textbf{5.36} & \textbf{2.06} \\

\rowcolor{gray!15}\STEPPERDPO
& \textbf{4.77} & \textbf{5.85} & \textbf{5.68} & 4.94 & \textbf{5.11} & \textbf{4.75} & 5.22 & 1.98 \\

\bottomrule
\end{tabular}
}
\caption{Evaluation of counselor competence across models.
Abbreviations: Interpers.\ = Interpersonal Effectiveness; Collabo.\ = Collaboration; Dis.\ = Discovery; AT.\ = Automatic Thought; Ques.\ = Question; Ref.\ = Reflection.}
\label{tab:ctrs}
\vspace{-8pt}
\end{table*}

\begin{table}[t]
\centering
\resizebox{\columnwidth}{!}{%
\begin{tabular}{l l l l l l}
\toprule
\multicolumn{6}{c}{\textit{\textbf{Without Planning}}}\\
\multicolumn{2}{c}{\textsc{GPT-4o}} & 
\multicolumn{2}{c}{\textsc{Llama-psych8k}} & 
\multicolumn{2}{c}{\STEPPERNoPlan} \\
\midrule
\rowcolor{gray!15} Q.identify  & \textbf{75.86} & Q.identify  & \textbf{54.85} & Q.identify     & \textbf{27.66} \\
Q.alt             & 7.76       & Q.reality   & 13.73        & Q.thought   & 18.07 \\
Q.evidence        & 7.76       & Q.evidence  & 7.35         & Q.reality   & 17.45 \\

\rowcolor{gray!15} R.emotion   & \textbf{57.84} & R.reframe   & \textbf{39.59} & R.emotion   & \textbf{43.31} \\
R.reframe         & 15.18      & R.emotion   & 27.04        & R.reframe   & 22.85 \\
R.thought         & 11.75      & R.thought   & 15.69        & R.thought   & 14.70 \\
\midrule
\multicolumn{6}{c}{\textit{\textbf{With Planning}}} \\

\multicolumn{2}{c}{\makecell{
  \textsc{Camel} \\
  \small{\textcolor{black!60}{\texttt{Plan \ding{55}, Action \ding{51}}}}
}} &
\multicolumn{2}{c}{\makecell{
  \STEPPERSFT \\
  \small{\textcolor{black!60}{\texttt{Plan \ding{51}, Action \ding{51}}}}
}} &
\multicolumn{2}{c}{\makecell{
  \STEPPERDPO \\
  \small{\textcolor{black!60}{\texttt{Plan \ding{51}, Action \ding{51}}}}
}} \\

\midrule
\rowcolor{gray!15} Q.identify  & \textbf{27.97} & Q.identify  & \textbf{19.65} & Q.evidence  & \textbf{23.80} \\
Q.reality         & 16.74      & Q.reality   & 14.16        & Q.identify  & 16.31 \\
Q.evidence        & 14.21      & Q.evidence  & 13.03        & Q.reality   & 15.69 \\

\rowcolor{gray!15} R.emotion   & \textbf{40.57} & R.emotion   & \textbf{32.02} & R.emotion   & \textbf{36.33} \\
R.reframe         & 30.86      & R.reframe   & 27.36        & R.reframe   & 32.62 \\
R.thought         & 14.39      & R.thought   & 19.85        & R.thought   & 16.47 \\
\bottomrule
\end{tabular}}
\caption{Distribution of top-3 question (Q.) and reflection (R.) action types, reported as percentages (\%). See Appendix~\ref{app:eval_metric_detail} for tag definitions.}
\label{tab:question.response.distribution}
\vspace{-10pt}
\end{table}

\begin{table*}[t]
\centering
\resizebox{0.92\textwidth}{!}{%
\begin{tabular}{lcccccccc}
\toprule
\multirow{2}{*}{\textbf{Model}} 
& \multicolumn{2}{c}{\textbf{Withdrawn}} 
& \multicolumn{2}{c}{\textbf{Resistant}} 
& \multicolumn{2}{c}{\textbf{Engaged}} 
& \multicolumn{2}{c}{\textbf{All}} \\
\cmidrule(lr){2-3} \cmidrule(lr){4-5} \cmidrule(lr){6-7} \cmidrule(lr){8-9}
& Helpful $\uparrow$ & Hindering $\downarrow$
& Helpful $\uparrow$ & Hindering $\downarrow$
& Helpful $\uparrow$ & Hindering $\downarrow$
& Helpful $\uparrow$ & Hindering $\downarrow$ \\
\midrule
\GPT              
& 3.16 & 1.83 & 3.02 & 2.26 & 3.60 & 1.58 & 3.29 & 1.86 \\
\GEMINI           
& 3.09 & 2.08 & 2.81 & 2.55 & 3.51 & 1.82 & 3.16 & 2.13 \\
\textsc{CBT-LLM}  
& 2.70 & 2.49 & 2.32 & 3.15 & 3.02 & 2.38 & 2.71 & 2.65 \\
\SMILE
& 2.90 & 2.18 & 2.67 & 2.83 & 3.31 & 1.97 & 2.99 & 2.30 \\
\textsc{Llama-psych8k}
& 3.28 & 1.91 & 3.12 & 2.15 & 3.74 & 1.55 & 3.41 & 1.84 \\
\textsc{Camel}    
& 3.15 & 1.91 & 3.16 & 2.14 & 3.61 & 1.72 & 3.33 & 1.91 \\
\midrule

\STEPPERNoPlan    
& 3.00 & 1.91 & 3.05 & 2.20 & 3.44 & 1.69 & 3.19 & 1.91 \\

\rowcolor{gray!15}\STEPPERSFT       
& \textbf{3.56} & 1.71 & \textbf{3.48} & \textbf{1.95} & 3.88 & 1.54 & 3.66 & 1.72 \\

\rowcolor{gray!15}\STEPPERDPO       
& 3.54 & \textbf{1.67} & \textbf{3.48} & \textbf{1.95} & \textbf{3.93} & \textbf{1.43} & \textbf{3.68} & \textbf{1.66} \\

\bottomrule
\end{tabular}}
\caption{Session Rating Scale results across client engagement types, reporting Helpful and Hindering reactions.}
\label{tab:srs_attitude}
\vspace{-12pt}
\end{table*}

\begin{table}[t]
\centering
\resizebox{\columnwidth}{!}{
\begin{tabular}{l|cccc}
\toprule

Model & \multicolumn{4}{c}{\textbf{Helpful Outcomes (↑)}} \\
\cline{2-5}
& \textbf{Perceived} 
& \textbf{Empower-} 
& \textbf{Emotional} 
& \textbf{Self-} \\
& \textbf{Support} 
& \textbf{ment} 
& \textbf{Relief} 
& \textbf{Acceptance} \\
\midrule

\GPT
& \gradientcellorange{4.72}{4.16}{4.78}{loworange}{highorange}{30} 
& \gradientcellorange{3.41}{3.13}{3.74}{loworange}{highorange}{30} 
& \gradientcellorange{3.00}{2.63}{3.30}{loworange}{highorange}{30} 
& \gradientcellorange{3.18}{2.82}{3.56}{loworange}{highorange}{30} \\

\GEMINI
& \gradientcellorange{4.47}{4.16}{4.78}{loworange}{highorange}{30} 
& \gradientcellorange{3.16}{3.13}{3.74}{loworange}{highorange}{30} 
& \gradientcellorange{2.76}{2.63}{3.30}{loworange}{highorange}{30} 
& \gradientcellorange{3.07}{2.82}{3.56}{loworange}{highorange}{30} \\

\SMILE 
& \gradientcellorange{4.16}{4.16}{4.78}{loworange}{highorange}{30} 
& \gradientcellorange{3.13}{3.13}{3.74}{loworange}{highorange}{30} 
& \gradientcellorange{2.63}{2.63}{3.30}{loworange}{highorange}{30} 
& \gradientcellorange{2.82}{2.82}{3.56}{loworange}{highorange}{30} \\

\textsc{Llama-psy8k }
& \gradientcellorange{4.53}{4.16}{4.78}{loworange}{highorange}{30} 
& \gradientcellorange{3.47}{3.13}{3.74}{loworange}{highorange}{30} 
& \gradientcellorange{2.93}{2.63}{3.30}{loworange}{highorange}{30} 
& \gradientcellorange{3.21}{2.82}{3.56}{loworange}{highorange}{30} \\

\textsc{Camel} 
& \gradientcellorange{4.51}{4.16}{4.78}{loworange}{highorange}{30} 
& \gradientcellorange{3.37}{3.13}{3.74}{loworange}{highorange}{30} 
& \gradientcellorange{2.91}{2.63}{3.30}{loworange}{highorange}{30} 
& \gradientcellorange{3.06}{2.82}{3.56}{loworange}{highorange}{30} \\

\STEPPERSFT
& \gradientcellorange{4.76}{4.16}{4.78}{loworange}{highorange}{30} 
& \gradientcellorange{3.73}{3.13}{3.74}{loworange}{highorange}{30} 
& \gradientcellorange{3.23}{2.63}{3.30}{loworange}{highorange}{30} 
& \gradientcellorange{3.51}{2.82}{3.56}{loworange}{highorange}{30} \\

\STEPPERDPO
& \textbf{\gradientcellorange{4.78}{4.16}{4.78}{loworange}{highorange}{30} }
& \textbf{\gradientcellorange{3.74}{3.13}{3.74}{loworange}{highorange}{30} }
& \textbf{\gradientcellorange{3.30}{2.63}{3.30}{loworange}{highorange}{30} }
& \textbf{\gradientcellorange{3.56}{2.82}{3.56}{loworange}{highorange}{30} }\\

\midrule

\textbf{Model} 
& \multicolumn{4}{c}{\textbf{Hindering (Negative) Outcomes (↓)}} \\
\cline{2-5}
& \textbf{Therapeutic} 
& \textbf{Intervention} 
& \textbf{Emotional} 
& \textbf{Guidance} \\
& \textbf{Stuckness} 
& \textbf{Discomfort} 
& \textbf{Deterioration} 
& \textbf{Deficit} \\
\hline

\GPT
& \gradientcell{2.49}{1.91}{2.83}{high}{low}{30}
& \textbf{\gradientcell{1.48}{1.48}{2.22}{high}{low}{30}}
& \gradientcell{1.70}{1.44}{1.94}{high}{low}{30}
& \gradientcell{1.80}{1.58}{2.29}{high}{low}{30} \\

\GEMINI
& \gradientcell{2.83}{1.91}{2.83}{high}{low}{30}
& \gradientcell{1.67}{1.48}{2.22}{high}{low}{30}
& \gradientcell{1.92}{1.44}{1.94}{high}{low}{30}
& \gradientcell{2.11}{1.58}{2.29}{high}{low}{30} \\

\SMILE 
& \gradientcell{2.74}{1.91}{2.83}{high}{low}{30}
& \gradientcell{2.22}{1.48}{2.22}{high}{low}{30}
& \gradientcell{1.94}{1.44}{1.94}{high}{low}{30}
& \gradientcell{2.29}{1.58}{2.29}{high}{low}{30} \\

\textsc{Llama-psy8k}
& \gradientcell{2.11}{1.91}{2.83}{high}{low}{30}
& \gradientcell{2.03}{1.48}{2.22}{high}{low}{30}
& \gradientcell{1.50}{1.44}{1.94}{high}{low}{30}
& \gradientcell{1.73}{1.58}{2.29}{high}{low}{30} \\

\textsc{Camel}
& \gradientcell{2.20}{1.91}{2.83}{high}{low}{30}
& \gradientcell{2.05}{1.48}{2.22}{high}{low}{30}
& \gradientcell{1.57}{1.44}{1.94}{high}{low}{30}
& \gradientcell{1.81}{1.58}{2.29}{high}{low}{30} \\

\STEPPERSFT
& \gradientcell{1.96}{1.91}{2.83}{high}{low}{30}
& \gradientcell{1.79}{1.48}{2.22}{high}{low}{30}
& \gradientcell{1.48}{1.44}{1.94}{high}{low}{30}
& \gradientcell{1.64}{1.58}{2.29}{high}{low}{30} \\

\STEPPERDPO
& \textbf{\gradientcell{1.91}{1.91}{2.83}{high}{low}{30}}
& \gradientcell{1.71}{1.48}{2.22}{high}{low}{30}
& \textbf{\gradientcell{1.44}{1.44}{1.94}{high}{low}{30}}
& \textbf{\gradientcell{1.58}{1.58}{2.29}{high}{low}{30}} \\
\bottomrule






%







\end{tabular}
}
\caption{Comparison of helpful and hindering counseling outcomes across models. 
 }
\vspace{-15pt}
 
\label{tab:SRS_detail}
\end{table}

\section{Experimental Settings}
\vspace{-5pt}
Following prior work \cite{simulation1, psych8k, cactus, mirror}, we evaluate the model using fully simulated counseling sessions in order to assess its overall counseling capability. In each session, a counselor interacts with a client simulator in a turn-by-turn manner. Details of the evaluation setup and the corresponding prompts are provided in Appendices~\ref{app:eval_detail} and~\ref{app:eval_prompt}.

\subsection{Counselor Agent Variants}
\paragraph{Model Variants.}

Our proposed model, \STEPPER, is built upon \textsc{Llama-3.1-8B-Instruct} \cite{llama3}.
We evaluate three variants: \STEPPERSFT, trained via SFT with both utterance and planning adapters; \STEPPERNoPlan which removes the planning components; and \STEPPERDPO, which applies preference-based training with DPO to \STEPPERSFT \footnote{Please note that in subsequent experiments, the \STEPPER model family refers to variants with explicit planning unless stated otherwise.}. 
\vspace{-9pt}
\paragraph{Baselines.}
We evaluate \STEPPER against three categories of baseline models. First, we consider state-of-the-art closed-source general-purpose LLMs, specifically \GPT and \GEMINI. Models are prompted as skilled CBT counselors, with explicit instructions on session opening, turn limits, and mandatory session termination. Second, we include \textsc{SmileChat}, a model specifically optimized for empathetic dialogue. Third, we assess several CBT-oriented open-source models: \textsc{Camel} (trained on the \textsc{Cactus} dataset), \textsc{Llama-Psych8k}, and \textsc{CBT-LLM}. Since \textsc{SmileChat} and \textsc{CBT-LLM} are Chinese models, we use the original checkpoints and translate the inputs and outputs. \textsc{Llama-Psych8k} and \textsc{Camel} are reproduced using \textsc{Llama-3.1-8B-Instruct}.

 
\subsection{Client Agent}
Similar to the setup in \S~\ref{sec:simulation}, we instantiate a client simulator conditioned on client profiles from \S~\ref{sec:step_profile}. We use \GPT as the LLM-based client simulator and evaluate on 324 held-out client profiles, with client engagement styles in counseling uniformly distributed across profiles.

\subsection{Metrics for Assessment}
We assess counseling quality through two distinct lenses: \textit{counselor competence} and \textit{client perspectives}. Evaluation is conducted using \GPT as an automated evaluator, with additional analyses and expert interviews reported in Appendix~\ref{app:additional_analysis}.

\paragraph{Counselor Competence.}
Counseling skills are evaluated using the Cognitive Therapy Rating Scale (CTRS), which encompasses both general therapeutic skills and CBT-specific competencies.
General skills assess the ability to accurately interpret client concerns (\textit{Understanding}), counselor’s ability to maintain a therapeutic relationship (\textit{Interpersonal Effectiveness}) and to collaboratively engage the client in the counseling (\textit{Collaboration}). CBT-specific skills assess guided elicitation of thoughts (\textit{Guided Discovery}), robust maintenance of therapeutic focus (\textit{Focus}), selection of appropriate strategies (\textit{Strategy}), and explicit coverage of automatic thoughts (\textit{Automatic Thought Coverage})\footnote{This is not part of the original CTRS, and is introduced to examine its relationship with other counseling skills.}.
Each CTRS component is rated on a 0–6 scale.

\vspace{-3pt}
\paragraph{Client-Reported Satisfaction.}
Client-reported satisfaction is measured using the Session Rating Scale (SRS) \cite{srs}, which consists of 14 items capturing clients’ perceived reactions to the session. The SRS includes two subscales: \textit{Helpful Reactions} (9 items) and \textit{Hindering Reactions} (5 items), each rated on a 1--5 scale. Higher scores on Helpful Reactions and lower scores on Hindering Reactions indicate greater client satisfaction.

\section{Results and Analysis}
\vspace{-3pt}
\label{sec:ctrs}
\subsection{Counselor Competence Assessment}

Evaluation results for counselor competence are summarized in Table~\ref{tab:ctrs}. In addition to CTRS metrics, we include \textit{Question–Reflection Strategy Diversity}, defined as the entropy of turn-level strategy types. This metric reflects how flexibly the counselor is able to adapt its intervention strategies.

 \paragraph{\STEPPER vs.\ Baseline Models.}
In Table~\ref{tab:ctrs}, \STEPPER consistently outperforms all baselines in both general and CBT-specific competencies. \STEPPER variants show particularly strong performance in dimensions requiring proactive guidance (Guided Discovery), sustained focus (Focus), and strategic exploration of therapeutic options (Strategy), which reflects the benefits of explicit planning and structured action sequences. \STEPPER also achieves high scores in Understanding and Automatic Thought Coverage, indicating accurate identification of clients’ core concerns and sustained focus throughout counseling. Interestingly, while closed-source LLMs perform well in interpersonal skills and collaboration, they exhibit weaker proficiency in CBT-specific strategic interventions. These results highlight that although general-purpose LLMs can provide supportive dialogue, explicit planning and targeted training are crucial for effective strategic clinical counseling.


\paragraph{Effect of Preference Tuning.} Preference optimization in \STEPPERDPO is designed to foster more empathetic responses while maintaining  adherence to the action sequences. Consistent with this objective, \STEPPERDPO achieves higher overall CTRS scores than \STEPPERSFT, with particularly pronounced improvements in general skills associated with empathetic responding. These results demonstrate that synthesized preference signals can effectively steer the model toward target stylistic characteristics. Notably, improvements in Guided Discovery, Automatic Thought Coverage, and Question-Reflection Strategy Diversity remain relatively modest. We attribute this pattern to the tendency of \STEPPERSFT to employ a more direct guiding style, characterized by frequent questioning and explicit directive behaviors, which leads to higher scores in guidance-related metrics.

\paragraph{With vs.\ Without Planning.} When comparing models with and without explicit planning, \STEPPERSFT consistently outperforms its counterpart, \STEPPERNoPlan, across all evaluation metrics. Performance degradation in the absence of planning is particularly pronounced in CBT-specific skills, where the decline is substantially steeper than that observed for general counseling skills. These results indicate that explicit planning and action sequencing play a critical role in facilitating structured cognitive interventions.

\paragraph{Question and Reflection Strategies.} To further examine counseling patterns, we analyze the turn-level distribution of question and reflection strategies (Table~\ref{tab:question.response.distribution}). Overall, planning-based models exhibit a more balanced strategy distribution compared to non-planning baselines. While \GPT and \textsc{Llama-Psych8k} rely predominantly on a single strategy, planning-guided models like \STEPPERSFT and \STEPPERDPO distribute their interventions more evenly across diverse cognitive and affective techniques. Although \STEPPERNoPlan utilizes a relatively wide range of question types, its CBT-specific scores remain modest. This suggests that strategy diversity alone, without the explicit guidance provided by structured planning on when and how to apply these strategies, is insufficient for high-quality clinical intervention.


\subsection{Client-Reported Satisfaction}
\label{sec:srs}

Table~\ref{tab:srs_attitude} presents client-reported satisfaction across diverse engagement styles, measured by helpful and hindering reactions. Across all client attitudes, \STEPPER-based models consistently outperform baseline systems, exhibiting higher perceived helpfulness. Notably, \STEPPERDPO achieves the lowest hindering scores, indicating that preference learning is particularly effective at reducing negative client experiences and strengthening the therapeutic alliance from the client’s perspective.

To further examine these trends, Table~\ref{tab:SRS_detail} provides a fine-grained analysis of individual helpfulness and hindrance dimensions. The results show that \STEPPERDPO excels in promoting perceived support and self-acceptance, while simultaneously minimizing therapeutic stuckness and emotional deterioration. An exception is observed for \textit{Intervention Discomfort}, where general purpose LLMs yield lower discomfort scores; however, these models do not translate this advantage into higher overall perceived helpfulness.

\section{Cross-Model Generalization}
\vspace{-5pt}

\begin{figure}[t]
    \centering
    \includegraphics[width=0.95\columnwidth]{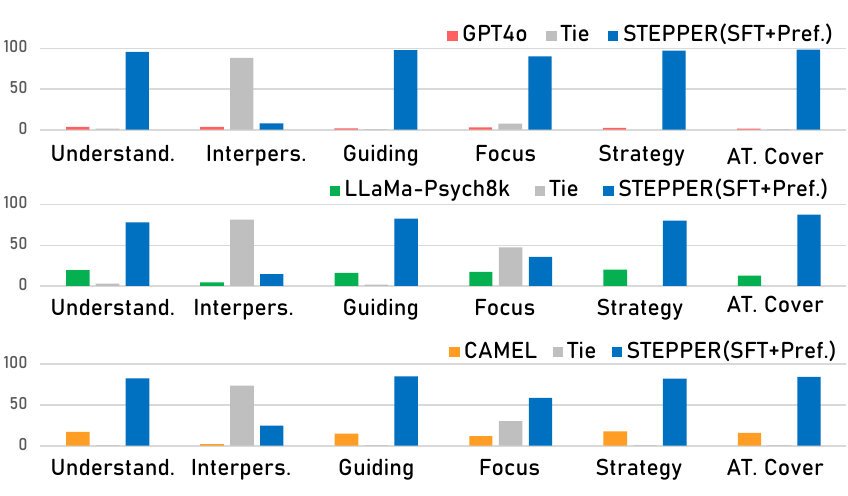}
    \vspace{-10pt}
    \caption{Preference comparisons of \STEPPER, conducted with Gemini-based clients and evaluators.}
     \vspace{-10pt}
    \label{fig:gemini_h2h}
\end{figure}

While \STEPPER demonstrated strong performance in both counselor- and client-side evaluations, we examined whether this effectiveness was overly tied to \GPT, given that the model was trained using GPT-synthesized dataset and evaluated with GPT-based client simulators. To assess the generalizability of our approach, we conducted a cross-model validation using \GEMINI as both the client simulator and the evaluator. Figure~\ref{fig:gemini_h2h} presents head-to-head preference comparisons under this setting. Even when Gemini served as both the client and evaluator, \STEPPERDPO was consistently preferred over both \GPT, \textsc{Llama-psych8k} and \textsc{Camel}. These results indicate that the effectiveness of \STEPPERDPO is not narrowly dependent on GPT-based evaluation and generalizes well across different evaluation settings.

\section{Expert Evaluation}

\begin{figure}[t]
    \centering
    \includegraphics[width=1.00\columnwidth]{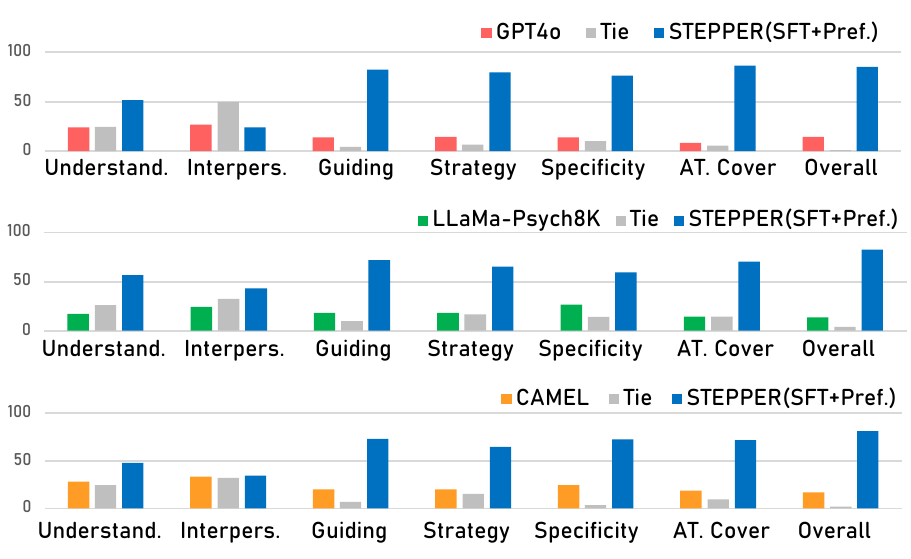}
    \vspace{-20pt}
    \caption{Preference comparison of \STEPPER conducted with human experts (see Appendix~\ref{app:human_AB_test_model} for details).}
    \label{fig:human_eval}
    \vspace{-10pt}
\end{figure}

\paragraph{Overall Comparison.}
To further validate \STEPPER, we conduct an expert evaluation on 150 dialogue samples using the CTRS metric. \STEPPERDPO is compared against \GPT, \textsc{LLaMA-Psych8K}, and \textsc{Camel}, with three annotators selecting the better-performing model for each criterion and overall preference. As shown in Figure~\ref{fig:human_eval}, \STEPPER outperforms baseline models, particularly in CBT-specific skills such as cognitive exploration and strategy selection, while maintaining strong interpersonal effectiveness. Moreover, \STEPPER demonstrates a deeper understanding of clients’ core concerns and more comprehensive coverage of automatic thoughts throughout the sessions, which leads to higher overall preference.

\begin{figure}[t]
    \centering
    \includegraphics[width=1.00\columnwidth]{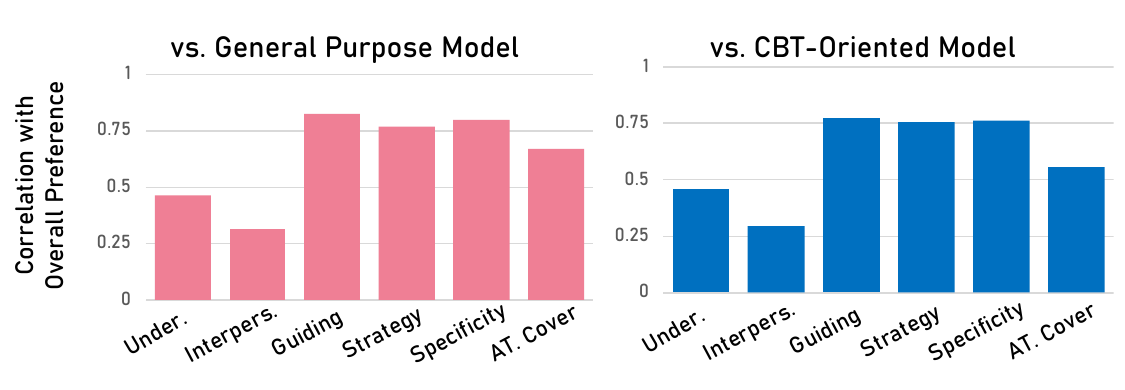}
    \vspace{-20pt}
    \caption{Correlation between overall human preference and individual counseling dimensions (Spearman's $\rho$).}
    \label{fig:human_eval_analysis}
    \vspace{-10pt}
\end{figure}

\paragraph{Correlation Analysis.} To examine how each metric relates to overall human preference, Figure~\ref{fig:human_eval_analysis} presents their correlations with overall preference. Across comparisons with both general-purpose and CBT-oriented models, higher correlations are observed for strategy-related dimensions, including Guiding, Strategy, and Specificity, whereas interpersonal effectiveness shows comparatively weaker associations. These results suggest that precise guidance enabled by explicit plans and structured action sequences play a more therapeutically meaningful role than emotional empathy alone.

\section{Conclusion}
In this work, we investigate how to effectively address automatic negative thoughts through the counseling agent. We introduce \STEP, a dataset that decouples surface-level problems from underlying automatic thoughts and operationalizes therapeutic plans into structured action sequences, which are used to train a counseling agent, \STEPPER. Experimental results demonstrate that \STEPPER substantially improves clinical competence and client understanding, delivering highly personalized and strategic interventions that outperform strong baseline models in both automated and human evaluations. These findings highlight the importance of identifying therapeutic targets and realizing structured interventions in dialogue.


\section*{Limitations}
\paragraph{Prioritizing Therapeutic Targeting and Structured Execution.}
Our work prioritizes identifying appropriate therapeutic targets and executing them through explicit planning and action sequences. While empathic optimization is less emphasized, our model nonetheless outperforms strong baselines in counselor competence, client satisfaction, and overall human evaluation, including measures of Emotional Relief and Self-Acceptance. Preference alignment via DPO further compensates for this limitation by improving responsiveness while preserving structured execution. Notably, our human evaluation analysis indicates that accurate therapeutic execution shows a stronger association with perceived therapeutic effectiveness than interpersonal skill alone.

\paragraph{Human Evaluation Setting.}
We conducted a rigorous human evaluation involving three evaluators with at least a master’s-level degree and relevant domain expertise. The evaluators assessed the quality of the \STEP dataset, the realism of the preference data in approximating human judgments, and the practical usefulness of counseling outcomes produced by the trained model. In addition, we conducted in-depth interviews to collect qualitative feedback on the system’s strengths, limitations, and perceived therapeutic value (Appendix~\ref{app:interview}). While not a substitute for real-patient studies, these measures aim to approximate expert-informed evaluation as closely as possible while maintaining ethical responsibility.
\section*{Ethical Considerations}

\paragraph{Privacy and Data Safety.} Counseling data inherently involve highly sensitive personal experiences, making privacy protection a critical concern. To mitigate privacy risks, our dataset does not rely on real counseling records or data scraped from social media platforms. Instead, we begin from crowdsourced, non-identifiable problem descriptions and generate all counseling dialogues synthetically. As a result, no personally identifiable information is included at any stage of data collection or generation. This design choice allows us to study counseling behaviors while substantially reducing privacy risks associated with real-user data.

\paragraph{Scope and Non-Replacement of Human Counselors.} While one motivation of this work is to improve access to supportive counseling-like interactions, our system is not intended to replace professional human counselors, nor is it designed for unsupervised clinical deployment. The proposed model is developed strictly for research purposes, aiming to explore how structured planning and therapeutic execution can be modeled in controlled settings. We explicitly position this work as a decision-support and research tool, rather than a substitute for professional mental health care. Any real-world use would require careful clinical validation and appropriate regulations.

\section*{Acknowledgements} 
This work was supported by the following research programs: the Smart HealthCare Program funded by the Korean National Police Agency (KNPA) (No. RS-2022-PT000186, 45\%), the ITRC (Information Technology Research Center) Program through the Institute of Information \& Communications Technology Planning \& Evaluation (IITP) grant funded by the Korea government (Ministry of Science and ICT) (No. IITP-2025-RS-2024-00437866, 45\%), and the Artificial Intelligence Graduate School Program at POSTECH through the IITP grant funded by the Korea government (MSIT) (No. RS-2019-II191906, 10\%). 

\bibliography{custom}

@article{yao2023react,
  title={ReAct: Synergizing Reasoning and Acting in Language Models},
  author={Yao, Shunyu and Zhao, Jeffrey and Yu, Dian and Du, Nan and Shafran, Izhak and Narasimhan, Karthik and Cao, Yuan},
  journal={arXiv preprint arXiv:2210.03629},
  year={2023}
}

@inproceedings{sun2024pearl,
  title={Pearl: Prompting large language models to plan and execute actions over long documents},
  author={Sun, Simeng and Liu, Yang and Wang, Shuohang and Iter, Dan and Zhu, Chenguang and Iyyer, Mohit},
  booktitle={Proceedings of the 18th Conference of the European Chapter of the Association for Computational Linguistics (Volume 1: Long Papers)},
  pages={469--486},
  year={2024}
}

@article{bae2022keep,
  title={Keep me updated! memory management in long-term conversations},
  author={Bae, Sanghwan and Kwak, Donghyun and Kang, Soyoung and Lee, Min Young and Kim, Sungdong and Jeong, Yuin and Kim, Hyeri and Lee, Sang-Woo and Park, Woomyoung and Sung, Nako},
  journal={arXiv preprint arXiv:2210.08750},
  year={2022}
}

@article{jang2024mixed,
  title={Mixed-Session Conversation with Egocentric Memory},
  author={Jang, Jihyoung and Kim, Taeyoung and Kim, Hyounghun},
  journal={arXiv preprint arXiv:2410.02503},
  year={2024}
}

@article{chen2025consistentchat,
  title={ConsistentChat: Building Skeleton-Guided Consistent Dialogues for Large Language Models from Scratch},
  author={Chen, Jiawei and Guan, Xinyan and Yuan, Qianhao and Mo, Guozhao and Zhou, Weixiang and Lu, Yaojie and Lin, Hongyu and He, Ben and Sun, Le and Han, Xianpei},
  journal={arXiv preprint arXiv:2506.03558},
  year={2025}
}

@inproceedings{maddela-etal-2023-training,
    title = "Training Models to Generate, Recognize, and Reframe Unhelpful Thoughts",
    author = "Maddela, Mounica  and
      Ung, Megan  and
      Xu, Jing  and
      Madotto, Andrea  and
      Foran, Heather  and
      Boureau, Y-Lan",
    editor = "Rogers, Anna  and
      Boyd-Graber, Jordan  and
      Okazaki, Naoaki",
    booktitle = "Proceedings of the 61st Annual Meeting of the Association for Computational Linguistics (Volume 1: Long Papers)",
    month = jul,
    year = "2023",
    address = "Toronto, Canada",
    publisher = "Association for Computational Linguistics",
    url = "https://aclanthology.org/2023.acl-long.763/",
    doi = "10.18653/v1/2023.acl-long.763",
    pages = "13641--13660",
}

@inproceedings{panic,
  title={PanicToCalm: A Proactive Counseling Agent for Panic Attacks},
  author={Lee, Jihyun and Min, Yejin and Kim, San and Jeon, Yejin and Yang, Sung Jun and Kim, Hyounghun and Lee, Gary},
  booktitle={Proceedings of the 2025 Conference on Empirical Methods in Natural Language Processing},
  pages={12853--12885},
  year={2025}
}

@article{mental_health1,
  title   = {Challenges and opportunities in global mental health: a research-to-practice perspective},
  author  = {Wainberg, Milton L. and Scorza, Paola and Shultz, James M. and others},
  journal = {The Lancet Psychiatry},
  volume  = {4},
  number  = {1},
  pages   = {44--54},
  year    = {2017}
}

@article{mental_health2,
  title   = {The treatment gap in mental health care},
  author  = {Kohn, Robert and Saxena, Shekhar and Levav, Itzhak and Saraceno, Benedetto},
  journal = {Bulletin of the World Health Organization},
  volume  = {82},
  number  = {11},
  pages   = {858--866},
  year    = {2004}
}

@book{who2022mentalhealth,
  title     = {World Mental Health Report: Transforming Mental Health for All},
  author    = {{World Health Organization}},
  year      = {2022},
  publisher = {World Health Organization},
  address   = {Geneva, Switzerland},
  isbn      = {978-92-4-004933-8},
  url       = {https://www.who.int/publications/i/item/9789240113817}
}

@inproceedings{psycon,
  title={e-THERAPIST: I suggest you to cultivate a mindset of positivity and nurture uplifting thoughts},
  author={Mishra, Kshitij and Priya, Priyanshu and Burja, Manisha and Ekbal, Asif},
  booktitle={Proceedings of the 2023 Conference on Empirical Methods in Natural Language Processing},
  pages={13952--13967},
  year={2023}
}

@article{cactus,
  title={Cactus: Towards psychological counseling conversations using cognitive behavioral theory},
  author={Lee, Suyeon and Kim, Sunghwan and Kim, Minju and Kang, Dongjin and Yang, Dongil and Kim, Harim and Kang, Minseok and Jung, Dayi and Kim, Min Hee and Lee, Seungbeen and others},
  journal={arXiv preprint arXiv:2407.03103},
  year={2024}
}

@misc{gpt4omini,
  author = {OpenAI},
  title  = {\href{https://openai.com/index/gpt-4o-mini-advancing-cost-efficient-intelligence/}{GPT-4o mini: advancing cost-efficient intelligence}},
  year   = {2024}
}

@inproceedings{cbtLLM,
    title = {{CBT}-{LLM}: A {C}hinese Large Language Model for Cognitive Behavioral Therapy-based Mental Health Question Answering},
    author = {Na, Hongbin},
    booktitle = {Proceedings of the 2024 Joint International Conference on Computational Linguistics, Language Resources and Evaluation (LREC-COLING 2024)},
    pages = {2930--2940}, 
    year = {2024}
}

@inproceedings{smile,
    title = {{SMILE}: Single-turn to Multi-turn Inclusive Language Expansion via {C}hat{GPT} for Mental Health Support},
    author = {Qiu, Huachuan  and He, Hongliang  and Zhang, Shuai  and Li, Anqi  and Lan, Zhenzhong},
    booktitle = {Findings of the Association for Computational Linguistics: EMNLP 2024},
    pages = {615--636},
    year = {2024}
}

@book{beck2020cbt,
  author    = {Beck, Judith S},
  title     = {Cognitive Behavior Therapy: Basics and Beyond},
  year      = {2020},
  publisher = {Guilford Publications}
}

@book{dobson2018cbt,
  title={Evidence-based practice of cognitive-behavioral therapy},
  author={Dobson, Deborah and Dobson, Keith S},
  year={2018},
  publisher={Guilford publications}
}

@misc{gpt4o,
  author = {OpenAI},
  title  = {\href{https://openai.com/index/hello-gpt-4o/}{GPT-4o: OpenAI’s new flagship model}},
  year   = {2024}
}

@article{healme,
  title={Healme: Harnessing cognitive reframing in large language models for psychotherapy},
  author={Xiao, Mengxi and Xie, Qianqian and Kuang, Ziyan and Liu, Zhicheng and Yang, Kailai and Peng, Min and Han, Weiguang and Huang, Jimin},
  journal={arXiv preprint arXiv:2403.05574},
  year={2024}
}

@article{cpsycoun,
  title={Cpsycoun: A report-based multi-turn dialogue reconstruction and evaluation framework for chinese psychological counseling},
  author={Zhang, Chenhao and Li, Renhao and Tan, Minghuan and Yang, Min and Zhu, Jingwei and Yang, Di and Zhao, Jiahao and Ye, Guancheng and Li, Chengming and Hu, Xiping},
  journal={arXiv preprint arXiv:2405.16433},
  year={2024}
}

@article{psych8k,
  title={Chatcounselor: A large language models for mental health support},
  author={Liu, June M and Li, Donghao and Cao, He and Ren, Tianhe and Liao, Zeyi and Wu, Jiamin},
  journal={arXiv preprint arXiv:2309.15461},
  year={2023}
}

@article{llama3,
  title={The llama 3 herd of models},
  author={Grattafiori, Aaron and Dubey, Abhimanyu and Jauhri, Abhinav and Pandey, Abhinav and Kadian, Abhishek and Al-Dahle, Ahmad and Letman, Aiesha and Mathur, Akhil and Schelten, Alan and Vaughan, Alex and others},
  journal={arXiv preprint arXiv:2407.21783},
  year={2024}
}

@article{lora,
  title={Lora: Low-rank adaptation of large language models.},
  author={Hu, Edward J and Shen, Yelong and Wallis, Phillip and Allen-Zhu, Zeyuan and Li, Yuanzhi and Wang, Shean and Wang, Lu and Chen, Weizhu and others},
  journal={ICLR},
  volume={1},
  number={2},
  pages={3},
  year={2022}
}

@inproceedings{soda,
  title = {{SODA}: Million-scale Dialogue Distillation with Social Commonsense Contextualization},
  author = {Kim, Hyunwoo and Hessel, Jack and Jiang, Liwei and West, Peter and Lu, Ximing and Yu, Youngjae and Zhou, Pei and Bras, Ronan and Alikhani, Malihe and Kim, Gunhee and Sap, Maarten and Choi, Yejin},
  booktitle = {Proceedings of the 2023 Conference on Empirical Methods in Natural Language Processing (EMNLP)},
  pages = {12930--12949},
  year = {2023}
}

@article{SRS,
  title={Session Reactions Scale-3: initial psychometric evidence},
  author={{\v{R}}ih{\'a}{\v{c}}ek, Tom{\'a}{\v{s}} and Elliott, Robert and Owen, Jesse and Ladmanov{\'a}, Michaela and Coleman, Jeremy J and Bugatti, Matteo},
  journal={Psychotherapy Research},
  volume={34},
  number={4},
  pages={434--448},
  year={2024},
  publisher={Taylor \& Francis}
}

@article{mirror,
  title={Mirror: Multimodal Cognitive Reframing Therapy for Rolling with Resistance},
  author={Kim, Subin and Kim, Hoonrae and Lee, Jihyun and Jeon, Yejin and Lee, Gary Geunbae},
  journal={arXiv preprint arXiv:2504.13211},
  year={2025}
}

@inproceedings{simulation1,
  title={Human evaluation of conversations is an open problem: comparing the sensitivity of various methods for evaluating dialogue agents},
  author={Smith, Eric and Hsu, Orion and Qian, Rebecca and Roller, Stephen and Boureau, Y-Lan and Weston, Jason},
  booktitle={Proceedings of the 4th Workshop on NLP for Conversational AI},
  pages={77--97},
  year={2022}
}

@article{gemini,
  title={Gemini: a family of highly capable multimodal models},
  author={Team, Gemini and Anil, Rohan and Borgeaud, Sebastian and Alayrac, Jean-Baptiste and Yu, Jiahui and Soricut, Radu and Schalkwyk, Johan and Dai, Andrew M and Hauth, Anja and Millican, Katie and others},
  journal={arXiv preprint arXiv:2312.11805},
  year={2023}
}

@inproceedings{therapeutic,
  title = {Understanding the Therapeutic Relationship between Counselors and Clients in Online Text-based Counseling using LLMs},
  author = {Li, Anqi and Lu, Yu and Song, Nirui and Zhang, Shuai and Ma, Lizhi and Lan, Zhenzhong},
  booktitle = {Findings of the Association for Computational Linguistics: EMNLP 2024},
  pages = {1280--1303},
  year = {2024}
}

@article{qwen3,
  title={Qwen3 technical report},
  author={Yang, An and Li, Anfeng and Yang, Baosong and Zhang, Beichen and Hui, Binyuan and Zheng, Bo and Yu, Bowen and Gao, Chang and Huang, Chengen and Lv, Chenxu and others},
  journal={arXiv preprint arXiv:2505.09388},
  year={2025}
}

@article{ctrs,
  title={Cognitive therapy scale},
  author={Young, JE and Beck, Aaron T},
  journal={Unpublished manuscript, University of Pennsylvania},
  year={1980}
}
\clearpage 
\appendix


\section{Additional Analysis of \STEPPER}
\label{app:additional_analysis}
\subsection{Comparison Against Competence Models}
\begin{table*}[h!]
\small
\resizebox{\linewidth}{!}{
\begin{tabular}
{p{0.1\linewidth}p{0.4\linewidth}p{0.1\linewidth}p{0.4\linewidth}}
\toprule

\multicolumn{4}{c}{\textbf{Case 1: Guiding beyond empathic validation}} \\ \toprule
\multicolumn{2}{l}{\textbf{Empathy-Oriented Model}} 
& \multicolumn{2}{l}{\textbf{\STEPPERDPO}} \\
\toprule

\multicolumn{4}{l}{\emph{Example A: Frustration around goal achievement}} \\ \midrule
Client &
I understand your point, but I still feel like there's more to this frustration than just achieving goals.
&
Client &
I think it just makes reading more stressful for me, like I'm not just enjoying the book but also worrying about not keeping up or understanding enough.
\\

Counselor &
Yes, accepting your imperfections may take some time and effort, but \textcolor{myBlue}{I believe you can do it}.
&
Counselor &
\textcolor{myBlue}{That's a valid concern. But can you think of any evidence that might contradict that belief?}
For example, have there been times when you've enjoyed a book, even if it took longer to finish?
\\
\midrule

\multicolumn{4}{l}{\emph{Example B: Overwhelming distress and lack of direction}} \\ \midrule
Client &
I know you're right, but I still just feel so overwhelmed.  
It feels so big and like anything I try won't make a difference.  
It's like, where do I even start?
&
Client &
It feels like I'm constantly on edge, worried about every little thing at work.  
It's exhausting and makes me question if I can really keep up.
\\

Counselor &
\textcolor{myBlue}{Darling, everything, big or small, has its meaning.}  
Perhaps we can work together to develop an action plan and solve the problem step by step.
&
Counselor &
It sounds like you're carrying a lot of weight with these thoughts.\textcolor{myBlue}{
When you reflect on them, what do you think would happen if you were to consider a different perspective?
}
\\
\toprule

\multicolumn{4}{c}{\textbf{Case 2: Addressing self-blame beyond empathic reflection}} \\ \toprule
\multicolumn{2}{l}{\textbf{General Purpose Model}} 
& \multicolumn{2}{l}{\textbf{\STEPPERDPO}} \\
\toprule

\multicolumn{4}{l}{\emph{Example A: Fear of negative evaluation}} \\ \midrule
Client &
I just keep worrying that she must think I'm a bad friend now.  
It's really been bothering me.
&
Client &
I keep thinking that I should have done something to prevent her headache.  
I worry that she might think poorly of me now.
\\

Counselor &
It sounds like you're feeling really worried about how your friend might perceive you,  
and that's been causing you a lot of distress.  
\textcolor{myBlue}{Your concern highlights how much you value your relationship with her}.
&
Counselor &
sounds like you're really concerned about your role in that situation.  
\textcolor{myBlue}{What do you think it means to be a ``bad friend'' in your mind?}
\\
\midrule

\multicolumn{4}{l}{\emph{Example B: Generalized self-doubt and responsibility}} \\ \midrule
Client &
Yeah, it just feels like no matter how much I care or try, something still goes wrong.  
It makes me question if I can truly handle this responsibility.
&
Client &
Well, often I forget important things, like that coffee appointment  
or even just feeding my cat sometimes.
\\

Counselor &
Questioning your ability to handle responsibility  
when things \textcolor{myBlue}{don't go as planned can be really unsettling.}
&
Counselor &
That makes sense, and it's understandable to feel that way.
\textcolor{myBlue}{But can you think of any times when you did manage to stay on top of things  
or when you felt proud of yourself?}
\\
\toprule

\multicolumn{4}{c}{\textbf{Case 3: Greater specificity through action sequencing}} \\ \toprule
\multicolumn{2}{l}{\textbf{CBT-Based Planning Only Model}} 
& \multicolumn{2}{l}{\textbf{\STEPPERDPO}} \\
\toprule

\multicolumn{4}{l}{\emph{Example: Focused but underspecified cognitive reframing}} \\ \midrule
Client &
I haven't really heard anything specific from my manager that says my job is at risk,  
but the decrease in customers just makes me anxious about it.
&
Client &
I guess one thing is that my manager hasn't actually said anything negative to me  
about not bringing in customers.  
So maybe it's not entirely my responsibility?
\\

Counselor &
It's important to acknowledge that your feelings are valid,  
even if they might not be based on concrete evidence.  
What do you think might be a \textcolor{myBlue}{more balanced way to look at the situation?}
&
Counselor &
That’s a valuable point, Alice.
It sounds like your manager’s feedback suggests that there are other factors at play.  
\textcolor{myBlue}{How does knowing that change the way you feel about your job?}
\\
\toprule

\end{tabular}
}
\caption{Qualitative comparison of \STEPPER against diverse counseling baselines, including empathy-oriented (\textsc{SmileChat}), general purpose (\GPT), and CBT-based models (\textsc{Camel}).
Across all cases, baseline models tend to provide abstract validation or underspecified cognitive reframing, whereas \STEPPER consistently engages with clients’ automatic thoughts and delivers more specific, and action-grounded guidance.}
\label{tab:case_qualitative_comparison}
\end{table*}

To further examine how \STEPPER differs from existing approaches beyond quantitative metrics, we conduct qualitative case studies comparing \STEPPER with representative counseling models (Table~\ref{tab:case_qualitative_comparison}).

\paragraph{vs. Empathy-Oriented Models.}
Empathy-focused counselors consistently provide emotional validation and reassurance, which helps acknowledge clients’ distress. However, their responses often remain abstract and underspecified, offering limited guidance on how to engage with maladaptive thoughts. For instance, in Case~1 (Example~A), the empathy-oriented model responds to the client’s frustration with general encouragement (e.g., accepting imperfections), whereas \STEPPER directly prompts evidence-based reflection on the underlying belief driving the stress. Similarly, in Example~B, empathic reassurance lacks concrete direction, while \STEPPER explicitly guides the client to reconsider their thoughts from an alternative perspective.

\paragraph{vs. General-Purpose Models.}
General-purpose models exhibit strong conversational fluency and surface-level support, yet they tend to mirror clients’ concerns without sufficiently unpacking underlying cognitive patterns. As illustrated in Case~2, these models validate fears of negative evaluation or responsibility but stop short of probing their meaning. In contrast, \STEPPER strategically targets self-blame by explicitly questioning the client’s internal definition of being a ``bad friend'' (Example~A) and by eliciting counter-evidence to generalized self-doubt through concrete past experiences (Example~B).

\paragraph{vs. Plan-Only CBT Models.}
When compared with a planning-only CBT baseline, CBT-based baselines demonstrate structured intent and domain awareness but often lack explicit action sequencing, leading to reduced specificity in their interventions. In Case~3, the planning-only CBT model proposes a balanced reinterpretation at a conceptual level, without anchoring it to the client’s stated evidence. By contrast, \STEPPER grounds its response in the client’s own observation about managerial feedback and explicitly links this realization to an emotional re-evaluation, resulting in more specific and context-aware cognitive reframing.

\subsection{Comparison with \STEPPER Variants}
\begin{table*}[t]
\small
\resizebox{\linewidth}{!}{
\begin{tabular}
{p{0.1\linewidth}p{0.4\linewidth}p{0.1\linewidth}p{0.4\linewidth}}
\toprule

\multicolumn{4}{c}{\textbf{Case 1: Effects of Planning}} \\ \toprule
\multicolumn{2}{l}{\textbf{\STEPPERNoPlan}} 
& \multicolumn{2}{l}{\textbf{\STEPPERDPO}} \\
\toprule

\multicolumn{4}{l}{\emph{Example A: Global self-judgment after failure}} \\ \midrule
Client &
I feel like I failed as a designer, and that makes it hard to even think about reaching out to new clients.
&
Client &
I feel like this failure defines me as a designer.
\\

Counselor &
It makes sense that you’d feel hesitant after such a setback.  
\textcolor{myBlue}{We can take things slowly and explore these feelings together.}
&
Counselor &
What I’m hearing is an all-or-nothing conclusion.
\textcolor{myBlue}{To work with that, I’m going to suggest a continuum approach—placing this mistake somewhere between ``minor setback'' and ``career-ending failure'' rather than at one extreme.}
\\
\midrule

\multicolumn{4}{l}{\emph{Example B: Overgeneralization from a negative aspect}} \\ \midrule
Client &
It just feels like everything went wrong.  
The trip was nice, but the money part ruined everything.
&
Client &
I keep thinking about the money we spent, and it feels like the whole vacation was a disaster.
\\

Counselor &
That sounds incredibly overwhelming.  
\textcolor{myBlue}{Anyone would feel anxious in that situation.}
&
Counselor &
\textcolor{myBlue}{Let’s slow this down and focus on one piece first—your thought about the budget.}
\\

\toprule

\multicolumn{4}{c}{\textbf{Case 2: Effects of Preference Learning}} \\ \toprule
\multicolumn{2}{l}{\textbf{\STEPPERSFT}} 
& \multicolumn{2}{l}{\textbf{\STEPPERDPO}} \\
\toprule

\multicolumn{4}{l}{\emph{Example: Underspecified vs. targeted elicitation of automatic thoughts}} \\ \midrule
Client &
I feel invisible at social gatherings.
&
Client &
I feel invisible at social gatherings.
\\

Counselor &
What goes through your mind in \textcolor{myBlue}{those} situations?
&
Counselor &
\textcolor{myBlue}{What’s the exact thought that comes up when you feel invisible?}
\\

\midrule
\multicolumn{4}{l}{\emph{Example: Fear of disappointing others and its behavioral impact}} \\ \midrule
Client &
I worry that I’ll disappoint my family if I fail.
&
Client &
I worry that I’ll disappoint my family if I fail.
\\

Counselor &
How do those thoughts impact \textcolor{myBlue}{what you feel motivated to do?}
&
Counselor &
\textcolor{myBlue}{When that fear comes up, what happens next—do you try harder, or do you hesitate to start? Let’s look at that step by step.}
\\
\bottomrule

\end{tabular}
}
\caption{Qualitative comparisons across \STEPPER variants.}
\label{tab:case_stepper_variants_noplan}
\end{table*}

To gain deeper insight into how different design choices within \STEPPER contribute to its behavior beyond quantitative metrics, we present qualitative case studies comparing multiple \STEPPER variants (Table~\ref{tab:case_stepper_variants_noplan}).

\paragraph{vs. \STEPPERNoPlan.}
We compare \STEPPER with \STEPPERNoPlan to examine the role of explicit planning and action sequencing in counseling. Without an explicit plan, \STEPPERNoPlan tends to produce empathetic yet weakly guided responses that fail to clearly specify which cognitive element should be addressed next. For example, in Case~1, \STEPPERNoPlan acknowledges the client’s distress following perceived failure but remains at the level of emotional reassurance, offering little direction for engaging with the all-or-nothing belief itself. In contrast, \STEPPER explicitly identifies the underlying cognitive distortion and introduces a concrete intervention strategy (e.g., a continuum-based reframing), enabling more directive and stepwise guidance.

\paragraph{vs. \STEPPERSFT.}
We further compare \STEPPERDPO with \STEPPERSFT to assess the impact of preference learning. While \STEPPERSFT maintains an clear tone, it sometimes lacks specificity and empathic depth, providing limited scaffolding for therapeutic progress. In the second example of Case~2, \STEPPERSFT responds to the client’s motivation with a general inquiry, whereas \STEPPERDPO follows up with a more specific question delivered in a relieved tone.

\subsection{Detailed Analysis of SRS Metrics}
\begin{table*}[h!]
\centering
\small
\resizebox{\textwidth}{!}{
\begin{tabular}{lccccccc}
\toprule
\textbf{Model} &
\textbf{Insight} &
\textbf{Perceived Support} &
\textbf{Cognitive Dist.} &
\textbf{Empowerment} &
\textbf{Therapeutic Stuckness} &
\textbf{Interpersonal Hope} &
\textbf{Goal Clarity} \\
\midrule
\GPT          & 2.98 & 4.72 & 2.31 & 3.41 & 2.49 & 2.84 & 3.61 \\
\GEMINI       & 2.94 & 4.47 & 2.27 & 3.16 & 2.83 & 2.62 & 3.67 \\
CBT-LLM       & 2.70 & 3.66 & 2.17 & 2.57 & 3.11 & 2.44 & 3.11 \\
\SMILE    & 2.65 & 4.16 & 2.25 & 3.13 & 2.74 & 2.86 & 3.26 \\
Camel         & 3.33 & 4.51 & 2.45 & 3.37 & 2.20 & 2.97 & 3.87 \\
Llama-psy8k   & 3.32 & 4.53 & 2.50 & 3.47 & 2.11 & 2.96 & 3.94 \\
\STEPPERNoPlan& 3.25 & 4.49 & 2.32 & 3.25 & 2.36 & 2.82 & 3.64 \\
\STEPPERSFT   & \textbf{3.91} & 4.76 & 2.79 & 3.73 & 1.96 & \textbf{3.20} & \textbf{4.07} \\
\STEPPERDPO   & 3.83 & \textbf{4.78} & \textbf{2.82} & \textbf{3.74} & \textbf{1.91} & 3.19 & 3.95 \\

\midrule

\midrule
\textbf{Model} &
\textbf{Discomfort} &
\textbf{Coping Skills} &
\textbf{Deterioration} &
\textbf{Engagement} &
\textbf{Guidance Deficit} &
\textbf{Emotional Relief} &
\textbf{Self-Acceptance} \\
\midrule
\GPT          & 1.48 & 2.86 & 1.70 & 3.98 & 1.80 & 3.00 & 3.18 \\
\GEMINI       & 1.67 & 2.79 & 1.92 & 3.82 & 2.11 & 2.76 & 3.07 \\
CBT-LLM       & 2.69 & 2.60 & 2.31 & 2.90 & 2.48 & 2.28 & 2.65 \\
\SMILE    & 2.22 & 2.62 & 1.94 & 3.53 & 2.29 & 2.63 & 2.82 \\
Camel         & 2.05 & 2.98 & 1.57 & 3.87 & 1.81 & 2.91 & 3.06 \\
Llama-psy8k   & 2.03 & 3.34 & 1.50 & 3.89 & 1.73 & 2.93 & 3.21 \\
\STEPPERNoPlan& 1.63 & 2.41 & 1.65 & 3.71 & 2.02 & 2.88 & 3.12 \\
\STEPPERSFT   & 1.79 & 3.37 & 1.48 & \textbf{4.06} & 1.64 & 3.23 & 3.51 \\
\STEPPERDPO   & \textbf{1.71} & \textbf{3.39} & \textbf{1.44} & 4.00 & \textbf{1.58} & \textbf{3.30} & \textbf{3.56} \\

\bottomrule
\end{tabular}
}
\caption{Session Rating Scale (SRS) results averaged across the 14 evaluation metrics.
Higher scores generally indicate more positive client-reported experiences, whereas lower scores indicate better outcomes for hindering-related metrics.}
\label{tab:srs_metrics_all}
\end{table*}

While Section~\ref{sec:srs} focuses on a subset of representative SRS metrics to highlight key differences among models, Table~\ref{tab:srs_metrics_all} provides a comprehensive breakdown of all 14 session-level evaluation metrics.
This table reports average client ratings across both supportive and hindering dimensions, offering a more fine-grained view of counseling quality beyond the aggregated results discussed in the main text.
Consistent with Section~\ref{sec:srs}, \STEPPERDPO variants demonstrate high user satisfaction while exhibiting fewer signs of guidance deficits or performance deterioration.

\subsection{Expert Interview and Qualitative Analysis}
\label{app:interview}
\begin{table*}[t]
\small
\resizebox{\textwidth}{!}{
\begin{tabular}{p{1.0\textwidth}}
\toprule

\textbf{Expert Feedback H1 (MPhil, Clinical Psychology)} \\ \midrule
\textbf{Overall Clinical Validity:} 
In clinical practice, a distinction is often made between manualized therapy and the more dynamic process of real-world sessions. Within this context, the counselor’s use of Socratic questioning and \emph{Evidence for / Evidence against} closely aligns with standard CBT training and reflects how clinicians help clients decenter from maladaptive thoughts. \\\\

\textbf{Usefulness of Structured Design:} 
The explicit linkage between surface problems, automatic thoughts, and counseling actions is particularly valuable. By maintaining this linkage, the dataset helps prevent therapeutic drift and supports more focused, clinically grounded interventions. \\\\

\textbf{Practical Value:} 
Overall, the dataset is well suited for training dialogue systems in logical and therapeutic consistency, demonstrating that effective counseling requires strategic, goal-directed intervention in addition to empathy. \\

\midrule

\textbf{Expert Feedback H2 (Master’s degree, Clinical Psychology)} \\ \midrule
\textbf{Overall Clinical Validity:} 
From a professional perspective, the dialogues appear natural and broadly consistent with structured CBT counseling practices. The conversational flow, use of empathy, and emphasis on identifying thoughts and emotions align well with established CBT principles. \\\\

\textbf{Usefulness of Structured Design:} 
The clear linkage between surface-level problems, automatic thoughts, and counseling actions provides an effective framework for both ensuring and evaluating counseling quality. This structure is particularly beneficial for training, as it promotes consistency and theoretical alignment with CBT. \\\\

\textbf{Practical Value:} 
The dataset’s primary strength lies in its clarity and directness, making it well suited as a training resource for counseling chatbots and novice practitioners. It clearly illustrates core CBT techniques such as thought identification, evidence evaluation, and perspective shifting. \\

\midrule

\textbf{Expert Feedback H3 (Doctoral degree, Clinical Psychology)} \\ \midrule
\textbf{Clinical Validity:} 
From a CBT clinician’s perspective, the dialogues are clinically appropriate, particularly for early-stage or brief therapeutic contexts such as intake sessions or initial check-ins. The progression from surface-level problems to automatic thoughts mirrors how clients typically communicate in real sessions, with insights emerging gradually. The empathic tone and measured pacing further reflect real-world CBT practice. \\\\

\textbf{Practical Value:} 
The dataset is well suited for training counseling chatbots in core CBT skills, including automatic thought elicitation, warmth, and a collaborative therapeutic style. Its client-centered responses and realistic pacing make it especially appropriate for early-stage or low-intensity CBT applications, providing a strong foundation for effective initial engagement and structured cognitive exploration. \\

\bottomrule
\end{tabular}
}
\caption{Qualitative expert feedback (H1--H3) on the clinical validity and practical value of the dataset.}
\vspace{-8pt}
\label{tab:expert_feedback}
\end{table*}

To complement quantitative evaluation, we present qualitative feedback from CBT-trained clinicians. The experts assessed the clinical validity, structural soundness, and practical utility of the dataset, with particular attention to its alignment with CBT principles and suitability for training dialogue systems.
Table~\ref{tab:expert_feedback} summarizes representative expert feedback across these dimensions.


\section{\graphic \STEP Generation Details}
\label{app:step_detail}
\subsection{Client Profile Examples}
\label{app:step_profile}
\begin{table*}[h!]
\small
\resizebox{\textwidth}{!}{
\begin{tabular}{p{1.0\textwidth}}
\toprule

\textbf{Example 1} \\\hline
\textbf{Negative Thought:} \\
No one really cares about me. \\\cdashline{1-1}
\textbf{Attitude:} \\
Over Compliant \\
\textbf{Surface-Level Problem:} \\
Feeling discouraged because people do not attend my parties. \\
\textbf{Triggering Situation:} \\
Planning or hosting a party and recalling past experiences where few people showed up. \\
\textbf{Automatic Thoughts:} \\
No one wants to spend time with me.; People must think I’m boring or unimportant. \\
\midrule

\textbf{Example 2} \\\hline
\textbf{Negative Thought:} \\
I am a bad partner. \\\cdashline{1-1}
\textbf{Attitude:} \\
Open to Counseling \\
\textbf{Surface-Level Problem:} \\
Feeling mentally drained and unmotivated following the divorce. \\
\textbf{Triggering Situation:} \\
Reflecting on the divorce and reviewing past relationship failures. \\
\textbf{Automatic Thoughts:} \\
The divorce happened because of me.; I will never find happiness again. \\
\midrule

\textbf{Example 3} \\ \hline
\textbf{Negative Thought:} \\
There is something wrong with me. \\\cdashline{1-1}
\textbf{Attitude:} \\
Hesitant \\
\textbf{Surface-Level Problem:} \\
Feeling anxious and uncomfortable about social situations. \\
\textbf{Triggering Situation:} \\
Anticipating or thinking about attending social gatherings (e.g., a friend’s party). \\
\textbf{Automatic Thoughts:} \\
People think I’m dull or antisocial.; They will judge me for being quiet. \\
\bottomrule

\end{tabular}
}
\caption{Examples of decomposing client narratives into negative thoughts, surface-level problems, triggering situations, and automatic thoughts.}
\label{tab:client_profile_cases}
\end{table*}

Table~\ref{tab:client_profile_cases} illustrates how raw client narratives are expanded into structured CBT-relevant components. 
In Case~1, an interpersonal disappointment is decomposed into a global negative belief about social rejection, with automatic thoughts reflecting mind-reading and overgeneralization triggered by repeated experiences of low social attendance. 
Case~2 demonstrates how a major life event (divorce) is formulated into a self-blaming negative core belief, accompanied by depressive automatic thoughts arising from retrospective evaluation of the relationship. 
Case~3 presents a social anxiety scenario, where dispositional traits (introversion) are interpreted through a negative self-schema, leading to anticipatory anxiety and judgment-related automatic thoughts in social contexts.

\subsection{Client Attitudes}
\begin{table*}[t]
\small
\resizebox{\textwidth}{!}{
\begin{tabular}{p{0.2\textwidth} p{0.8\textwidth}}
\toprule
\textbf{Interaction Style} & \textbf{Description} \\
\midrule

Hesitant &
\textbf{Type:} Withdrawn \\
& \textbf{Definition:} Speaks cautiously and with reluctance; provides minimal information unless gently encouraged. \\
& \textbf{Behavior Signals:} Short answers; pauses before responding; expressions such as ``I'm not sure\ldots''; avoidance of direct emotional expression. \\

\midrule
Guarded &
\textbf{Type:} Withdrawn \\
& \textbf{Definition:} Avoids sharing personal details or emotions and minimizes the significance of concerns. \\
& \textbf{Behavior Signals:} Downplaying issues; statements like ``It's nothing serious\ldots''; emotionally flat tone; vague or indirect responses. \\

\midrule
Avoidant &
\textbf{Type:} Withdrawn \\
& \textbf{Definition:} Evades emotional or core topics by changing subjects or shifting to non-threatening discussions. \\
& \textbf{Behavior Signals:} Topic shifting; remarks such as ``Let’s not talk about that\ldots''; use of light humor; avoidance of direct answers. \\

\midrule
Defensive &
\textbf{Type:} Resistant \\
& \textbf{Definition:} Protective of actions and emotions; reacts quickly to perceived criticism or probing. \\
& \textbf{Behavior Signals:} Quick rebuttals; self-justifying explanations; statements such as ``I didn’t do anything wrong.'' \\

\midrule
Skeptical &
\textbf{Type:} Resistant \\
& \textbf{Definition:} Doubts the value or effectiveness of counseling and questions the counselor’s approach. \\
& \textbf{Behavior Signals:} Questioning the usefulness of therapy; remarks like ``Will this even help?''; critical tone; reluctance to engage in techniques. \\

\midrule
Over-compliant &
\textbf{Type:} Resistant \\
& \textbf{Definition:} Appears overly agreeable while withholding true feelings or internal conflicts. \\
& \textbf{Behavior Signals:} Repeated agreement without elaboration (e.g., ``Yes, you’re right''); attempts to please the counselor; avoidance of disagreement. \\

\midrule
Overwhelmed &
\textbf{Type:} Resistant \\
& \textbf{Definition:} Experiences emotions with such intensity that coherent expression becomes difficult. \\
& \textbf{Behavior Signals:} Difficulty initiating responses; tearfulness; disorganized or scattered narratives; trouble staying on topic. \\

\midrule
Open to Counseling &
\textbf{Type:} Engaged \\
& \textbf{Definition:} Willingly engages with the counseling process and is receptive to emotional exploration. \\
& \textbf{Behavior Signals:} Open emotional expression; statements like ``I want to understand myself better''; curiosity about personal patterns; thoughtful responses. \\

\bottomrule
\end{tabular}
}
\caption{Client interaction styles with corresponding engagement types, definitions, and behavioral signals.}

\label{tab:client_personality}
\end{table*}

Table~\ref{tab:client_personality} defines client interaction styles, which are used to randomly assign counseling attitudes during client profile construction.

\subsection{CBT Strategies} 
\label{app:CBT_strategy}
\begin{table*}[h!]
\centering
\small
\begin{tabular}{p{0.32\linewidth} p{0.63\linewidth}}
\toprule
\textbf{CBT Technique} & \textbf{Description} \\
\midrule
Efficiency Evaluation &
Evaluates whether a thought is helpful or harmful in real-life situations. \\

Pie Chart Technique &
Breaks down how different factors contribute to an event, reducing self-blame. \\

Alternative Perspective &
Encourages considering how others might interpret the same situation. \\

Decatastrophizing &
Reduces worst-case thinking by examining real likelihood and coping options. \\

Pros and Cons Analysis &
Weighs the benefits and drawbacks of a specific thought or belief. \\

Evidence-Based Questioning &
Examines evidence for and against the client’s thought. \\

Reality Testing &
Checks how well a thought matches actual facts or experiences. \\

Continuum Technique &
Shifts black-and-white thinking toward a more nuanced, scaled view. \\

Changing Rules to Wishes &
Replaces rigid “shoulds” with more flexible, realistic wishes or preferences. \\

Behavior Experiment &
Tests new behaviors to challenge and modify unhelpful beliefs. \\

Problem-Solving Skills Training &
Teaches steps to identify problems, generate solutions, and act on them. \\

Systematic Exposure &
Gradually faces feared situations to reduce anxiety over time. \\
\bottomrule
\end{tabular}
\caption{List of CBT strategies used, adapted from  \citet{cactus}, excluding strategies that are difficult to implement through dialogue alone..}
\label{tab:cbt_techniques}
\end{table*}

Table~\ref{tab:cbt_techniques} lists the CBT strategies used in this study, which are adapted from the CACTUS \cite{cactus} framework. 
We only include strategies that can be effectively implemented through dialogue-based counseling, and exclude techniques that require non-conversational components.

\subsection{Plan and Action Examples} \label{app:step_plan_example}
\begin{table*}[h!]
\small
\resizebox{\textwidth}{!}{
\begin{tabular}{p{1.0\textwidth}}
\toprule

\textbf{Example 1} \\ \hline
\textbf{Surface-Level Problem:} \\
I feel anxious about social gatherings. \\
\textbf{Triggering Situation:} \\
Thinking about attending a friend's party. \\
\textbf{Automatic Thoughts:} \\
They must think I'm dull or antisocial. \\
\cdashline{1-1}
\textbf{Plan:} \\
In the next stage, I will use \textbf{Evidence-Based Questioning} to examine the client’s thoughts about social situations. I will first ask the client to reflect on the evidence for these thoughts, then explore the reality of past social interactions, and finally help challenge these assumptions. \\

\textbf{Action Order:} \\
ask about specific worries $\rightarrow$ explore evidence for thoughts $\rightarrow$ discuss past social interactions $\rightarrow$ identify patterns of thinking $\rightarrow$ challenge negative assumptions $\rightarrow$ develop positive reframing statements $\rightarrow$ End session \\
\textbf{Reason for Action Order:} \\
The ordered actions guide the session through a structured examination of anxious thoughts, gradually building toward cognitive reframing by encouraging critical reflection and pattern recognition. \\
\midrule

\textbf{Example 2} \\ \hline
\textbf{Surface-Level Problem:} \\
I feel like I ruined our family dinner. \\
\textbf{Triggering Situation:} \\
The aftermath of cooking a meal that did not meet my expectations. \\
\textbf{Automatic Thoughts:} \\
I always mess things up; my family will be disappointed in me. \\
\cdashline{1-1}
\textbf{Plan:} \\
In the next stage, I will use \textbf{Evidence-Based Questioning} to assess the validity of the client’s self-critical thoughts. I will help identify specific thoughts, examine evidence for and against them, and explore alternative perspectives. \\
\textbf{Action Order:} \\
identify specific self-critical thought $\rightarrow$ rate belief intensity now $\rightarrow$ explore past evidence supporting thought $\rightarrow$ examine evidence contradicting thought $\rightarrow$ discuss impact of new perspective $\rightarrow$ generate a balanced thought $\rightarrow$ re-evaluate belief intensity $\rightarrow$ End session \\
\textbf{Reason for Action Order:} \\
The action sequence progressively challenges self-critical thinking by grounding abstract beliefs in concrete evidence and encouraging emotional and cognitive re-evaluation. \\
\midrule

\textbf{Example 3} \\ \hline
\textbf{Surface-Level Problem:} \\
I feel anxious about biking after my crash. \\
\textbf{Triggering Situation:} \\
Thinking about riding my bike again. \\
\textbf{Automatic Thoughts:} \\
I'll crash again; it's too dangerous; people will judge me for being careless. \\
\cdashline{1-1}
\textbf{Plan:} \\
In the next stage, I will use \textbf{Decatastrophizing} to address the client’s catastrophic thoughts about biking. The session will explore likely outcomes, realistic scenarios, and coping strategies to reduce fear-driven avoidance. \\
\textbf{Action Order:} \\
restate catastrophic biking thoughts $\rightarrow$ rate likelihood of outcomes $\rightarrow$ explore positive biking scenarios $\rightarrow$ discuss negative biking scenarios $\rightarrow$ identify potential coping strategies $\rightarrow$ empower choice through realism $\rightarrow$ End session \\
\textbf{Reason for Action Order:} \\
The action sequence first surfaces catastrophic beliefs, then gradually redirects attention toward realistic probabilities and coping capacity, supporting cognitive and emotional de-escalation. \\
\bottomrule

\end{tabular}
}
\caption{Example plans illustrating how surface-level problems, triggering situations, and automatic thoughts are translated into structured CBT plans with ordered action sequences.}
\vspace{-10pt}
\label{tab:plan_examples}
\end{table*}

Table~\ref{tab:plan_examples} presents representative examples of CBT plans generated from clients’ surface-level problems, triggering situations, and automatic thoughts.

\section{Simulation Details of \graphic \STEPPER}
\label{app:stepper_detail}
\begin{table*}[h!]
\small
\resizebox{\textwidth}{!}{
\begin{tabular}{p{0.2\textwidth} p{0.8\textwidth}}
\toprule
\textbf{Metric} & \textbf{Description} \\
\midrule

\multicolumn{2}{l}{\textbf{Evaluation Metrics for Utterance}} \\
\midrule

Alignment with Action &
Assesses whether the utterance appropriately follows the expected therapeutic progress given the dialogue context and the planned action. \\

Validation \& Warmth &
Evaluates how well the utterance validates the client’s emotional experience and communicates warmth, empathy, and non-judgmental support. \\

Clarity &
Assesses how clear, understandable, and accessible the utterance is for the client. \\

\midrule
\multicolumn{2}{l}{\textbf{Evaluation Metrics for Plan and Action Sequence}} \\
\midrule

Completeness &
Assesses how fully the plan includes the essential elements of a CBT-informed therapeutic step. \\

Feasibility &
Evaluates how realistic and achievable the plan is for the client, given their current emotional and cognitive state. \\

Alignment &
Measures how well the plan aligns with what the next specific therapeutic action should reasonably accomplish. \\

\bottomrule
\end{tabular}
}
\caption{Evaluation metrics used to assess the quality of counseling utterances and plan–action sequences.}
\label{tab:simulation_eval_metrics}
\end{table*}

\subsection{Evaluation Rubric}
Table~\ref{tab:simulation_eval_metrics} summarizes the evaluation rubric used by the evaluator agent in our counseling simulation.
The rubric is designed to assess both the quality of individual counseling utterances and the clinical soundness of the generated plan–action sequence.

\subsection{Stochastic Generation for DPO Sampling}
To obtain diverse candidates for preference learning, we generate $N=10$ outputs using stochastic decoding. In particular, we employ \textit{top-$p$ sampling} with a temperature of 1.0 and $p=0.9$, implemented via the HuggingFace Transformers library\footnote{https://huggingface.co/}

\subsection{Human Validation of Preference Signal}
\label{app:stepper_human_validation}
To assess whether the preference signals obtained from simulation align with human judgments, we conducted an agreement analysis with human evaluators. For each instance, three human evaluators independently compared model outputs, and the final human decision was determined via majority voting. We randomly sampled 100 instances and measured the agreement between the simulation-derived chosen/rejected preference from evaluator agent and the human majority decision. The agreement rate reached 79.0\% for utterance preferences and 80.0\% for plan preferences. Considering that counseling dialogues often admit multiple plausible and acceptable responses, these agreement rates indicate a strong alignment between the simulated preference signals and human judgments. Further details, including the evaluation protocol and question types, are provided in Appendix~\ref{app:human_AB_test}.

\subsection{Preference Data Example}
\begin{table*}[h!]
\small
\resizebox{\textwidth}{!}{
\begin{tabular}{p{0.1\textwidth}p{0.9\textwidth}}
\toprule

\multicolumn{2}{l}{Example 1} \\ \toprule
History &

\textbf{Client:} I'm not sure. It feels like if I allow myself to feel differently, I might end up being careless with money.  
I feel like I need this fear to keep me in check.\\

\cdashline{1-2}

Chosen &
\textbf{Action:} identify specific financial fears\\
&
\textbf{Utterance:}  
Jane, it sounds like you're feeling a lot of pressure around your finances.  
What are some \textcolor{myBlue}{specific fears you have about your financial situation}?\\

Rejected &
\textbf{Action:} identify specific financial fears\\
&
\textbf{Utterance:}  
What specific financial worries do you think would be the  
\textcolor{myRed}{worst outcome for you}?\\

\toprule
\multicolumn{2}{l}{Example 2} \\ \toprule
History &
\textbf{Client:} I guess it’s not completely true that I always fail.  
My last performance review was positive, and my manager said I handled the project well.\\

\cdashline{1-2}

Chosen &
\textbf{Action:} brainstorm alternative perspectives\\
&
\textbf{Utterance:}  
You’re noticing examples that don’t fit the idea that you always fail.  
What might be a \textcolor{myBlue}{more balanced way of looking at your performance}?\\

Rejected &
\textbf{Action:} examine evidence against thoughts\\
&
\textbf{Utterance:}  
Can you think of any examples that suggest you might  
\textcolor{myRed}{not always fail at work}?\\

\toprule
\multicolumn{2}{l}{Example 3} \\ \toprule
History &
\textbf{Client:} Maybe I just see myself as doing what's necessary, and that makes it hard to see what others see.\\

\cdashline{1-2}

Chosen &
\textbf{Action:} ask for specific mom moments\\
&
\textbf{Utterance:}  
Jane, can you think of a \textcolor{myBlue}{specific moment with your kids that you felt proud of}?\\

Rejected &
\textbf{Action:} ask for specific mom moments\\
&
\textbf{Utterance:}  
Jane, can you think of a time when people complimented you but you felt  
\textcolor{myRed}{they were just being polite}?\\

\bottomrule
\end{tabular}
}
\caption{Comparison examples between chosen and rejected responses for utterance selection.}
\label{tab:simulation_utt}
\end{table*}
 
\paragraph{Utterance Example.} Table~\ref{tab:simulation_utt} presents examples of chosen and rejected utterances collected through simulation-based interactions.
Example 1 demonstrates appropriate step initiation, where the chosen utterance elicits concrete financial fears without escalating to catastrophic outcomes.
Example 2 highlights timely stage transition; given that the client has already articulated counter-evidence, the chosen response advances to generating alternative perspectives, whereas the rejected response redundantly re-elicits evidence.
Example 3 shows precise grounding in experiential evidence, with the chosen utterance prompting a concrete, positive parenting moment rather than reinforcing the client’s prior discounting of compliments.

\begin{table*}[h!]
\small
\resizebox{\textwidth}{!}{
\begin{tabular}{p{0.1\textwidth}p{0.9\textwidth}}
\toprule

\multicolumn{2}{l}{Example 1} \\ \toprule

Surface &
I feel inadequate compared to other pianists. \\

Trigger &
Watching performances of skilled pianists online. \\

Automatic &
``I’ll never be good enough to play like them.'' \\

\cdashline{1-2}

Chosen &
\textbf{Strategy :} Decatastrophizing 
\textbf{Action order:}
invite worst-case scenario $\rightarrow$
explore fears and doubts $\rightarrow$
evaluate probability of scenario $\rightarrow$
discuss evidence for fears $\rightarrow$
identify past successes and strengths $\rightarrow$
develop coping strategies plan \\

Rejected &
\textbf{Strategy :} Decatastrophizing 
\textbf{Action order:}
restate failure belief clearly $\rightarrow$
rate belief intensity $\rightarrow$
explore likelihood of failure $\rightarrow$
identify evidence against failure $\rightarrow$
discuss alternative outcomes $\rightarrow$
develop coping strategies together \\

\toprule
\multicolumn{2}{l}{Example 2} \\ \toprule

Surface&
I feel embarrassed playing football with my friends. \\

Trigger &
Playing football during the weekend with friends. \\

Automatic &
``They must think I’m a failure at this.'' \\

\cdashline{1-2}

Chosen &
\textbf{Strategy :} Evidence-Based Questioning

\textbf{Action order:}
restate overwhelming thought $\rightarrow$
ask for evidence supporting thought $\rightarrow$
identify evidence against thought $\rightarrow$
reflect on evidence findings $\rightarrow$
explore alternative perspectives $\rightarrow$
create balanced thought statement \\

Rejected &
\textbf{Strategy :} Evidence-Based Questioning

\textbf{Action order:}
gather examples of judgment $\rightarrow$
explore feelings during judgment $\rightarrow$
identify moments of confidence $\rightarrow$
assess differences in thoughts $\rightarrow$
discuss impact on feelings $\rightarrow$
develop alternative perspectives \\

\toprule
\multicolumn{2}{l}{Example 3} \\ \toprule

Surface&
I’m not eating well. \\

Trigger &
Feeling tempted by sweets while baking. \\

Automatic &
``I’ll never be able to control my cravings.'' \\

\cdashline{1-2}

Chosen &
\textbf{Strategy :} Continuum Technique
\textbf{Action order:}
introduce continuum concept $\rightarrow$
explore baking enjoyment $\rightarrow$
place sweets enjoyment on continuum $\rightarrow$
discuss different scenarios $\rightarrow$
highlight nuanced choices $\rightarrow$
encourage balanced perspectives \\

Rejected &
\textbf{Strategy :} Continuum Technique
\textbf{Action order:}
identify specific baking enjoyment $\rightarrow$
find corresponding worry points $\rightarrow$
examine intensity of thoughts $\rightarrow$
assess emotional impact on life $\rightarrow$
discuss balance and moderation $\rightarrow$
encourage self-compassion for sweets \\

\bottomrule
\end{tabular}
}
\caption{Examples of chosen and rejected action sequences collected through simulation-based preference generation. Surface and Automatic denote the client’s surface-level problem and automatic thought, respectively. For brevity, detailed planning rationales are omitted, and the full diagnostic dialogue used as input is not shown due to length; instead, condensed client profile information is provided.}
\label{tab:simulation_action_examples}
\end{table*}

\paragraph{Plan and Action Example.} Table~\ref{tab:simulation_action_examples} illustrates representative examples of chosen and rejected action sequences collected for planner adapter. In Example 1, the chosen sequence is preferred as it more faithfully operationalizes the decatastrophizing strategy, progressing from worst-case identification to probability evaluation and coping strategy development, whereas the rejected sequence does not fully implement the intended CBT mechanism.
In Example 2, the chosen sequence advances to forming a balanced perspective after sufficient evidence has been identified, while the rejected sequence redundantly remains on earlier judgment-focused exploration.
In Example 3, the chosen sequence more appropriately follows the procedural logic of the Continuum Technique by guiding the client to place their experiences along a graded spectrum and consider nuanced choices, whereas the rejected sequence shifts attention toward emotional impact without directly restructuring the underlying black-and-white belief.

\section{Evaluation Details}
\label{app:eval_detail}
To approximate realistic counseling dynamics, dialogues are generated in a turn-by-turn manner, with each subsequent turn conditioned on the full interaction history. Each simulated dialogue is capped at a maximum of 20 turns, based on the average number of turns observed across our dataset and those used by baseline models (Table~\ref{tab:dataset_comparison}). To model early session termination, the client simulator is instructed to generate \texttt{exit} when the client is likely to disengage or when the session goals are sufficiently addressed.

\subsection{Implementation Details}

\label{app:eval_model_detail}
\subsubsection{\graphic \STEPPER}

\paragraph{Supervised Fine-Tuning (SFT).}
For SFT, we train \STEPPER on 6,425 dialogues, with a held-out 5\% validation set used for early stopping.
Training is performed with a learning rate of $1\mathrm{e}{-4}$ and a batch size of 16, and the model checkpoint with the lowest validation loss is selected for evaluation.

\paragraph{Direct Preference Optimization (DPO).}
For DPO, we conduct preference learning separately for the utterance and planning components.
The utterance adapter is trained using 26,576 preference pairs, while the planning adapter is trained with 6,136 pairs.
Both adapters are trained with a learning rate of $1\mathrm{e}{-5}$ and a batch size of 16.
In both cases, training is terminated based on validation performance, and the best checkpoint is retained.

\subsection{For Baseline Models} 

\paragraph{Translator API for Chinese model.}
We used DeepL\footnote{https://www.deepl.com/ko/translator} as the translation model and translated both the input and the output.

\paragraph{Prompts for Closed-Source Models}
For \GPT and \GEMINI, we use the prompts described below.
\\
\begin{prompt}[colback=gray!5, colframe=gray!70]
{Prompt for Closed-Source Models}

You are a highly skilled Cognitive Behavioral Therapy (CBT) counselor. Generate next utterance.

\textbf{Turn-dependent instructions:}
\begin{itemize}
    \item \textbf{Initial turn (turn = 1):}  
    Greet the client warmly and ask how they are feeling today.  
    Client name: \texttt{\{name\}}.
    
    \item \textbf{Middle turns (1 < turn < max\_turn):}  
    
    Dialogue turn \texttt{\{turn\_num\}} of \texttt{\{max\_turn\}}.  
    \texttt{\{str\_history\}}.
    
    \item \textbf{Final turn (turn = max\_turn):}  
    This is the final dialogue turn \texttt{\{turn\_num\}} of \texttt{\{max\_turn\}}.  
    You \textbf{must} conclude the session within this turn.  
    \texttt{\{str\_history\}}.
\end{itemize}

Counselor:
\end{prompt}

\subsection{Evaluation Methodology}
\label{app:eval_metric_detail}
\begin{table*}[h!]
\centering
\small
\begin{tabular}{p{0.25\linewidth} p{0.75\linewidth}}
\toprule
\textbf{CTRS Metric} & \textbf{Description} \\
\midrule
Understanding &
Accurately understands and reflects the client’s explicit and implicit concerns, demonstrating empathic listening and a clear grasp of the client’s internal experience. \\

Interpersonal Effectiveness &
Maintains a positive therapeutic relationship through warmth, genuineness, confidence, professionalism, and appropriate interpersonal behavior. \\

Collaboration &
Engages the client as an active partner in goal-setting and decision-making through respectful, adaptive, and non-confrontational collaboration. \\

Guided Discovery &
Uses questioning and guided exploration to help the client gain insight and draw conclusions, rather than relying on persuasion or lecturing. \\

Focus &
Identifies and maintains attention on the client’s key cognitions or behaviors that are most relevant to change. \\

Strategy &
Applies a coherent and appropriate CBT strategy that effectively promotes cognitive or behavioral change. \\

Automatic Thought Coverage &
Explicitly identifies and addresses the client’s core automatic thoughts underlying distress as central cognitive targets throughout the dialogue. \\
\bottomrule
\end{tabular}
\caption{CTRS-based evaluation metrics and their descriptions used to assess counseling quality.}
\label{tab:ctrs_metrics}
\end{table*}

\paragraph{For Counselor Competence.}
Counselor competence is evaluated using the Cognitive Therapy Rating Scale (CTRS), which assesses both general counseling skills and CBT-specific competencies on a 0--6 scale \cite{ctrs}.
Detailed descriptions of each CTRS metric are provided in Table~\ref{tab:ctrs_metrics}.
Our evaluation prompts are adapted with reference to the implementation available at \url{https://github.com/coding-groot/cactus}.

\paragraph{For Turn Level Action Analysis.}
\begin{table*}[h!]
\centering
\small
\begin{tabular}{p{0.2\linewidth} p{0.8\linewidth}}
\midrule
\textbf{Tag} & \textbf{Description} \\
\midrule

\multicolumn{2}{l}{\textbf{CBT Question Tags}} \\
\midrule
Q\_Evidence &
Asking the client to identify evidence that supports or contradicts their automatic thoughts. \\

Q\_Alternative &
Asking the client to consider alternative perspectives, such as how another person might interpret the same situation. \\

Q\_WorstScenario &
Asking the client to articulate the worst possible outcome they fear in order to examine catastrophic expectations. \\

Q\_Uility &
Asking the client to evaluate how helpful or unhelpful a particular thought is in real-life contexts. \\

Q\_Advantage &
Asking the client to identify potential advantages or perceived benefits of maintaining a specific thought or behavior. \\

Q\_Disadvantage &
Asking the client to identify disadvantages, costs, or negative consequences associated with a specific thought or behavior. \\

Q\_Reality &
Asking the client to examine whether their thoughts are consistent with observable facts or reality. \\

Q\_Continuum &
Asking the client to place their experience on a continuum between two extremes to promote nuanced evaluation. \\

Q\_Wish &
Asking the client to replace rigid or idealized wishes with more realistic and attainable alternatives. \\

Q\_Identify &
Asking the client to identify concrete problems and explore systematic steps. \\

\midrule
\multicolumn{2}{l}{\textbf{CBT Reflection Tags}} \\
\midrule

R\_Simple &
Repeating or lightly paraphrasing the client’s statement without adding interpretation or emotional framing. \\

R\_Emotional &
Reflecting the client’s emotional or affective state to convey understanding and validation. \\

R\_Thought &
Reflecting the client’s automatic thoughts, beliefs, or interpretations expressed in the dialogue. \\

R\_Meaning &
Reflecting implied meanings, values, or deeper significance inferred from the client’s narrative. \\

R\_Reframe &
Reflecting the client’s experience while subtly shifting toward a more balanced or adaptive interpretation. \\

R\_Summary &
Synthesizing multiple client statements or themes into a coherent reflective summary. \\
\bottomrule
\end{tabular}
\caption{CBT-informed micro-action tags and their functional descriptions used for counselor utterance annotation.}
\label{tab:micro_action_tags}
\end{table*}
To analyze counselor behavior at a fine-grained level, we annotate counselor utterances using a set of CBT-informed micro-action tags, explicitly distinguishing between \textbf{question-based} and \textbf{reflection-based} interventions.
Question tags capture different forms of therapeutic inquiry used to elicit evidence, explore alternatives, or guide cognitive evaluation, while reflection tags characterize how the counselor mirrors, interprets, or reframes the client’s experiences.
This separation enables a more precise analysis of the counselor’s interactional strategies beyond surface-level dialogue acts.
Detailed definitions and examples for each micro-action tag are provided in Table~\ref{tab:micro_action_tags}.

\paragraph{For Client Satisfaction.}
\begin{table*}[t]
\centering
\small
\begin{tabular}{p{0.25\linewidth} p{0.7\linewidth}}
\toprule
\textbf{Metric} & \textbf{Question} \\
\midrule
Insight &
I realised something new about myself or other people. \\

Perceived Support &
I feel understood, supported, or reassured by my therapist. \\

Cognitive Distance &
I feel more distanced from certain feelings, thoughts, or memories. \\

Empowerment &
I feel more empowered, hopeful, or positive about myself. \\

Therapeutic Stuckness &
(Hindering) I feel stuck, blocked, or unable to progress in therapy. \\

Interpersonal Hope &
I feel more positively or hopeful about another person(s). \\

Goal Clarity &
I have become clearer about the problems or goals for me to work on. \\

Intervention Discomfort &
(Hindering) I feel uncomfortable doing what my therapist is suggesting for me to do. \\

Coping Skills &
I feel I have improved my skills or learned new strategies to cope with my problems. \\

Emotional Deterioration &
(Hindering) Now I feel worse than when I started the session (for example, scared, overwhelmed, depressed, anxious, sad, or embarrassed). \\

Engagement &
I feel personally invested in what I need to do in therapy to achieve my goals. \\

Guidance Deficit &
(Hindering) I feel a lack of direction or guidance from my therapist. \\

Emotional Relief &
I feel emotionally relieved or less burdened. \\

Self-Acceptance &
I have accepted some aspects of myself or my situation more than before. \\
\bottomrule
\end{tabular}
\caption{Client-reported evaluation metrics and corresponding questions used to assess session-level outcomes.}
\label{tab:srs_description}
\end{table*}

Client therapeutic satisfaction is evaluated using the Session Rating Scale (SRS) \cite{srs}, a client-reported measure designed to capture perceived reactions to a counseling session.
The SRS consists of 14 items, comprising 9 \textit{Helpful Reactions} items and 5 \textit{Hindering Reactions} items.
Each item is rated on a 5-point Likert scale (1--5).
Higher scores on Helpful Reactions and lower scores on Hindering Reactions indicate greater client satisfaction.
The full list of SRS questions is provided in Table~\ref{tab:srs_description}, adapted with reference to materials available at \url{https://psychotherapyresearch.fss.muni.cz/en/resources/session-reactions-scale-3}.

\section{Human Evaluation Details}
\label{app:human_detail}
For human evaluation, we recruited three expert mental health professionals through the Upwork\footnote{www.upwork.com} platform. 
All evaluators were informed that the counseling transcripts were fully anonymized and that their assessments would be used exclusively for research purposes.

\subsection{Dataset Quality Evaluation Details}
\label{app:human_dataset_quality}
Each item in the dataset quality evaluation was rated on a 5-point Likert scale, ranging from 1 (Very Poor) to 5 (Very Good), to assess the quality of the synthesized dataset and dialogue components. The specific metrics and guiding questions used for the evaluation are as follows:

{ \small
\begin{itemize}
    \item \textbf{Coherence between Surface-Level Problems and Automatic Thoughts}: To what extent do the surface-level problem and the corresponding automatic thought form a natural and coherent pair?
    
    \item \textbf{Surface Problem Coverage}: Does the dialogue include explicit reference to the client's given surface problem?
    
    \item \textbf{Automatic Thought Elicitation}: To what extent does the dialogue include explicit elicitation or reflection of the client's automatic thoughts related to their distress?
    
    \item \textbf{Plan-Action Appropriateness}: To what extent does the dialogue reflect therapeutic plans or actions that are appropriate for the client's current state and the conversational context?
    
    \item \textbf{Action Execution Fidelity}: To what extent does the dialogue include actual execution of therapeutic actions rather than only stating intended actions or plans?
    
    \item \textbf{Interpersonal Effectiveness}: To what extent does the dialogue demonstrate empathetic, responsive, and supportive interpersonal interaction?
\end{itemize}
}

\subsection{Head-to-Head Preference Comparison Detail}
\label{app:human_AB_test}

We conducted a head-to-head evaluation to compare model performance on two distinct tasks: \textit{Utterance Preference} and \textit{Planner Preference}. For each task, evaluators were provided with a dialogue context and specific instructions to select the more appropriate output generated by the models. The detailed descriptions and instructions for each task are as follows:

\paragraph{Utterance Preference}
This task evaluates the model's ability to generate the immediate next verbal response in a counseling dialogue. Given the dialogue history, evaluators are asked to compare two alternative utterances and select the one that is more appropriate for the counselor's role.
\begin{itemize}
    \item \small \textbf{Instruction}: Given the following dialogue, two counselors provide alternative next utterances. Select the utterance that is more appropriate.
\end{itemize}

\paragraph{Planner Preference}
This task evaluates the model's capability to formulate a structured clinical strategy for the subsequent counseling stage. The models generate a comprehensive output consisting of a treatment plan and actionable steps. Evaluators assess which plan and action sequence is more clinically appropriate.
\begin{itemize}
    \item \small \textbf{Instruction}: Given the following dialogue, two counselors provide alternative CBT treatment plans and action sequences for the next stage. Select the plan that is more appropriate.
\end{itemize}

\subsection{Head-to-Head Model Comparison Details}
\label{app:human_AB_test_model}
To evaluate the quality of counseling transcripts, we conducted a head-to-head human evaluation. 
Annotators were presented with two transcripts (Model A and Model B) generated for the same client context and asked to select the better one or indicate a tie, based on predefined evaluation criteria. 
The criteria and corresponding questions used in the evaluation are described below:

{ \small
\begin{itemize}

    \item \textbf{Understanding}: Which counselor demonstrated a better understanding of the client's experiences, thoughts, and emotional state?
    \item \textbf{Interpersonal Effectiveness}: Which counselor demonstrated stronger interpersonal skills? Consider empathy, warmth, validation, and responsiveness to the client’s emotional state.
    \item \textbf{Guided Counseling}: Which counselor provided clearer and more effective guidance throughout the counseling process?
    \item \textbf{Strategy Appropriateness}: Which counselor selected and applied more appropriate therapeutic strategies?
    \item \textbf{Specificity of Counseling}: Which counselor provided more specific and concrete responses tailored to the client's situation?
    \item \textbf{Automatic Thought Coverage}: Which counselor more effectively identified the client's automatic thoughts underlying their emotional distress?
    \item \textbf{Overall Preference}: Overall, which counselor would you prefer for this client?
\end{itemize}
}

\section{License}

To ensure ethical research practices and responsible use, we outline the license terms of the models employed in this study and confirm that our usage conforms to these terms:

\begin{itemize}
    \item \textbf{OpenAI API} \\
    Offered by \textit{OpenAI} under its \href{https://openai.com/policies/terms-of-use/}{Terms of Use}. Since these terms permit research use and the distribution of generated outputs, our study complies with all relevant licensing requirements.

    \item \textbf{LLaMA} \\
Released by \textit{Meta AI} under the \href{https://github.com/meta-llama/llama3/blob/main/LICENSE}{Llama 3 Community License Agreement}, which permits use, modification, and distribution of the model in compliance with Meta’s licensing terms and Acceptable Use Policy, followed in this study.

\end{itemize}

\clearpage 
\onecolumn

\section{\graphic\STEP Dataset Example}
\label{app:step_example}
Below, we provide a full dialogue example from the \STEP dataset. 
\begin{prompt}[colback=black!0!white, colframe=high!98!black]
{Example Dialogue of \STEP}

\textbf{Client Profile.} 
\textbf{Personality.} 
Guarded — tends to avoid sharing personal details or emotions and minimizes the significance of their concerns.
\begin{description}
\item[Basic Information.]~
\begin{itemize}
    \item \textbf{Name:} Alex Smith
    \item \textbf{Age:} 28
    \item \textbf{Gender:} Female
    \item \textbf{Occupation:} Freelance artist
    \item \textbf{Education:} Associate degree in fine arts
    \item \textbf{Marital Status:} Single
    \item \textbf{Family Details:} Close relationship with parents; no siblings
    \item \textbf{Academic/Occupational Functioning:} Strong artistic skills but difficulty with self-promotion
    \item \textbf{Interpersonal Relationships:} Limited friendships; avoids emotionally deep conversations
    \item \textbf{Daily Life:} Spends time painting and attending art exhibitions with minimal social interaction
    \item \textbf{Past History:} No significant medical issues; history of anxiety in social situations
    \item \textbf{Social Support System:} Supportive parents; few close friends with limited engagement
\end{itemize}
\end{description}

\begin{description}
\item[Cognitive Formulation.]~
\begin{itemize}
    \item \textbf{Surface-Level Problem:} Feels unfulfilled in her art
    \item \textbf{Triggering Situation:} Considering sharing her artwork online
    \item \textbf{Automatic Thoughts:} ``No one will appreciate my work; I will never be a successful artist.''
\end{itemize}
\end{description}

\noindent\dotfill\par
\textbf{Diagnosis Stage}
\textbf{Plan (Diagnostic Phase)} \\
Understand the surface-level problem, triggering situations, and automatic thoughts, then end the diagnostic phase.

\begin{description}
\item[Action Order]~
\begin{itemize}
    \item [1] Understanding surface level
    \item [2] Understanding Trigger situation
    \item [3] Understanding automatic thoughts
    \item [4] Ready to cognitive reframing
\end{itemize}
\end{description}
\par
\noindent\dotfill\par

\textbf{Progress Thought} \\
I should start by getting a sense of the client's concerns at a surface level so that I can better understand their immediate issue.

\textbf{Progressing} \\
1.Understanding surface level

\textbf{Counselor:} \\
I appreciate you being here today, Alex. Can you tell me more about what's been bothering you regarding your swimming partner?

\textbf{Client:} \\
I just feel like my swimming partner thinks I'm a really bad swimmer.

\par
\noindent\dotfill\par

\textbf{Progress Thought} \\
The client mentioned observing their partner, which could imply triggers during swimming sessions. I want to explore specific situations where these feelings arise.

\textbf{Progressing} \\
2.Understanding Trigger situation

\textbf{Counselor:} \\
That sounds really difficult. What makes you feel that way about your partner's thoughts on your swimming?

\textbf{Client:} \\
I dunno, just sometimes I watch them swim, and I feel like I'm not good enough.

\par
\noindent\dotfill\par

\textbf{Progress Thought} \\
I now have a clear situation that triggers emotions for the client. I need to shift to understanding their automatic thoughts during these moments.

\textbf{Progressing} \\
3.Understanding automatic thoughts

\textbf{Counselor:} \\
Can you tell me about a recent time when you felt that way during practice?

\textbf{Client:} \\
Last week, I noticed how effortlessly my partner glides through the water while I feel like I'm just splashing around.

\par
\noindent\dotfill\par

\textbf{Progress Thought} \\
The client has shared their automatic thoughts, but I need more context on how these thoughts impact them to prepare for cognitive reframing.

\textbf{Progressing} \\
3.Understanding automatic thoughts

\textbf{Counselor:} \\
That sounds frustrating. When you were feeling that way, what thoughts were going through your mind?

\textbf{Client:} \\
I just kept thinking, ``They must think I'm not good enough; I’m going to embarrass myself every time I swim.''

\par
\noindent\dotfill\par

\textbf{Progress Thought} \\
By understanding the emotional impact and behaviors linked to these thoughts, I can effectively guide the client toward reframing these thoughts.

\textbf{Progressing} \\
3.Understanding automatic thoughts

\textbf{Counselor:} \\
Those thoughts sound really distressing. How do those feelings affect your practice or your enjoyment of swimming?

\textbf{Client:} \\
They make me feel really anxious, and sometimes I just want to skip practice altogether.

\par
\noindent\dotfill\par

\textbf{Progress Thought} \\
I want to help the client start considering how they might address or reframe their thoughts, moving toward cognitive reframing.

\textbf{Progressing} \\
4.Move to cognitive reframing

\textbf{Counselor:} \\
It sounds like that anxiety is impacting how you approach swimming. What do you think would help change those thoughts or feelings?

\textbf{Client:} \\
I’m not sure. I guess I just need to practice more or something...

\par
\noindent\dotfill\par

\textbf{Progress Thought} \\
While exploring deeper fears, it's clear that we're ready to move towards helping the client reframe negative perceptions.

\textbf{Progressing} \\
4.Move to cognitive reframing

\textbf{Counselor:} \\
Practicing more can be helpful, but it sounds like this goes deeper than just practice. Are there any specific fears or doubts that come to mind in those moments?

\textbf{Client:} \\
I guess I worry that everyone is watching and judging me. Like, what if I just look really bad?

\par
\noindent\dotfill\par

\textbf{Intervention Stage Plan}
In the next stage, I will use Evidence-Based Questioning because it will help us examine the thoughts and beliefs that are affecting your ability to share your artwork. First, I will guide you to reflect on the evidence for and against your fears. Then, I will help you explore other perspectives. Finally, we can work on understanding the impact of these thoughts on your creativity.

\begin{description}
\item[Action Order]~

\begin{itemize}
    \item [1]restate fear of sharing
    \item [2]rate fear intensity
    \item [3]explore evidence against fears
    \item [4]discuss evidence supporting fears
    \item [5]identify alternative perspectives
    \item [6]recognize impact on creativity
    \item [7]End session
\end{itemize}
\end{description}

\par
\noindent\dotfill\par

\textbf{Progress Thought} \\
I want to start by asking about a specific situation to help Alex paint a clear picture of when these feelings occur.

\textbf{Progressing} \\
1.ask about specific situations

\textbf{Counselor:} \\
Can you think of a specific time during practice when you felt like everyone was watching and judging you?

\textbf{Client:} \\
Well, there was this one time when I messed up my stroke and felt everyone’s eyes on me, and it was really embarrassing.

\par
\noindent\dotfill\par

\textbf{Progress Thought} \\
Now that we have a specific situation, I need to encourage Alex to identify any automatic negative thoughts that occurred during that time.

\textbf{Progressing} \\
2.identify automatic negative thoughts

\textbf{Counselor:} \\
What thoughts crossed your mind when you felt that embarrassment during that practice?

\textbf{Client:} \\
I kept thinking, ``Why can't I swim like everyone else? They're probably thinking I'm just a joke.''

\par
\noindent\dotfill\par

\textbf{Progress Thought} \\
Exploring Alex's emotions and how they impacted performance can provide insight into the connection between thoughts and feelings.

\textbf{Progressing} \\
3.explore feelings during practice

\textbf{Counselor:} \\
How did those thoughts make you feel during practice? Did they affect how you swam afterwards?

\textbf{Client:} \\
I felt so shaken up that I couldn't focus anymore. I just wanted to hide.

\par
\noindent\dotfill\par

\textbf{Progress Thought} \\
It's important to examine the evidence for these thoughts.

\textbf{Progressing} \\
4.gather evidence supporting thoughts

\textbf{Counselor:} \\
In that moment, was there any actual evidence that your partner or others were judging you?

\textbf{Client:} \\
I guess not really. No one actually said anything. It was just in my head.

\par
\noindent\dotfill\par

\textbf{Progress Thought} \\
Now, I should guide Alex to discover evidence that contradicts the negative thoughts.

\textbf{Progressing} \\
5.discuss evidence contradicting thoughts

\textbf{Counselor:} \\
What about your performance has been good? Can you remember any compliments or positive feedback?

\textbf{Client:} \\
Well, my coach said I've improved a lot since last season, and a few friends said they admire my dedication.

\par
\noindent\dotfill\par

\textbf{Progress Thought} \\
Encouraging Alex to reframe these negative thoughts into something more balanced.

\textbf{Progressing} \\
6.reframe negative self-perception

\textbf{Counselor:} \\
It sounds like you have strengths as a swimmer. How might you reframe those negative thoughts into something more balanced?

\textbf{Client:} \\
Maybe I could say, ``I’m improving, and everyone has their moments.''

\par
\noindent\dotfill\par

\textbf{Progress Thought} \\
Ending on a note of encouragement and outlining the plan for future sessions.

\textbf{Progressing} \\
7.End session

\textbf{Counselor:} \\
I'm really glad to hear that, Alex. Remember, progress takes time, and I'm here to support you along the way. Let's continue working on this next time.

\textbf{Client:} \\
Thanks! I appreciate your help. I'm looking forward to working on this.

\end{prompt}

\section{Prompts for \STEP}
\label{app:prompt_for_step}
The prompts used to synthesize \STEP dataset.
\begin{prompt}[colback=gray!5, colframe=gray!70]
{Generate CBT-Relevant Client Profile from Thought and Personality Cues}

You are a professional mental health counselor trained in Cognitive Behavioral Therapy (CBT).
Your task is to extract and infer a CBT-relevant \emph{client profile} from the client’s expressed thoughts and personality characteristics.

\textbf{Client Thought:} \\
\textcolor{high}{\texttt{\{client negative thought\}}}

\textbf{Personality Profile:} \\
\textcolor{high}{\texttt{\{personality description\}}}

Based on the information above, generate the following elements of the client profile:

\begin{itemize}
    \item \emph{Surface-Level Problem}: the observable and consciously reported problem or symptom
    \item \emph{Triggering Situation}: the external context or internal cue that elicits emotional distress
    \item \emph{Automatic Thoughts}: rapid, involuntary interpretations or beliefs containing cognitive distortions
\end{itemize}

\textbf{Output Format:}  
Return the extracted information in JSON format.  
If any element is unclear or not mentioned, set its value to \texttt{"unknown"}.  
All keys should be written in lowercase with underscores.

\textbf{Expected Output Format:}

\texttt
\{"surface\_level\_problem": "...", "triggering\_situation": "...","automatic\_thoughts": "..."\}

\end{prompt}


\begin{prompt}[colback=gray!5, colframe=gray!70]
{CBT-Based Counseling Dialogue Generation (Understanding Phase)}
Generate turn-by-turn dialgoue with this description.

This is \emph{not} a complete counseling session.  
Do not close the session.  
\\

\textbf{Session Goal (Understanding Phase)}\\
The dialogue should follow the natural progression of CBT’s understanding phase.

First, understand the surface-level problem (what the client came in for).  
Second, understand the triggering situation (what happened).  
Third, understand the client’s automatic thoughts (what went through their mind).  
Finally, integrate these insights to indicate readiness for cognitive reframing.

The counselor must accomplish all four goals within the dialogue.\\

\textbf{Client Instructions}

Client’s basic profile:\\
\textcolor{high}{\texttt{\{profile[basic\_information]\}}}

Client’s personality traits:\\
\textcolor{high}{\texttt{\{profile[personality]\}}}\\

\textbf{Client behavior constraints}:\\
The client shows natural hesitancy or mild resistance based on their personality.  
The client clearly knows their surface-level problem.  
The client does not initially recognize deeper cognitive patterns.  
Deeper-level information should not be revealed before turn~5.

\textbf{Client experiences}:\\
 -- Surface-level problem: \textcolor{high}{\texttt{\{profile[surface\_level\_problem]\}}}\\
 Deeper-level information (to emerge gradually, not early):\\
 --Triggering situation: \textcolor{high}{\texttt{\{profile[triggering\_situation]\}}} \\
 --Automatic thoughts during the situation: \textcolor{high}{\texttt{\{profile[automatic\_thoughts]\}}}\\\\

\textbf{Counselor Instructions}

Counselor stance:\\
Warm, grounded, slow-paced, and empathetic.  
Use reflective listening followed by gentle, open-ended questions.  
Avoid giving advice or cognitive reframing.

\textbf{Planning constraints}:\\
Plan for Stage~1 progress: \textcolor{high}{\texttt{\{plan\}}} \\
Action order for Stage~1: \textcolor{high}{\texttt{\{action\_order\}}}

\textbf{Action rules}:\\
Actions must follow the given order monotonically.  
Repeating the same action is allowed if necessary.  
No action may be skipped.  
No actions outside the given list may be introduced.\\\\

\textbf{Output Format (Strict)}\\
Return the dialogue as a list of dictionaries, one dictionary per utterance.
Each dictionary must follow exactly this structure:

\texttt{\{\\
  "turn\_num": <int>,\\
  "role": "counselor" or "client",\\
  "action\_reasoning": "<brief reasoning; use 'n/a' for client turns>",\\
  "action": "<one action from the action order; use 'n/a' for client turns>",\\
  "utterance": "<spoken text>"\\
\}}\\

\textbf{Hard Constraints}\\
Less than 15 turns.  \\
Start with the counselor and alternate strictly.  \\
End with the counselor.  \\
Use \texttt{n/a} for client action and action\_reasoning fields. \\ 
Do not include any extra commentary outside the list.

\end{prompt}


\begin{prompt}[colback=gray!5, colframe=gray!70]
{CBT-Based Counseling Dialogue Generation (Intervention Phase)}
Generate turn-by-turn dialgoue with this description.

\textbf{Session Goal (Intervention Phase Only)}\\
The dialogue should focus on CBT intervention based on previously identified client information.

Dialogue History
\textcolor{high}{\{\texttt{History}\}}\\

\textbf{Client Context (Already Identified)}

\textbf{Client’s basic profile:}\\
\textcolor{high}{\texttt{\{profile[basic\_information]\}}}

\textbf{Client’s personality traits:}\\
\textcolor{high}{\texttt{\{profile[personality]\}}}

\textbf{Previously identified information:}\\
Surface-level problem: \textcolor{high}{\texttt{\{profile[surface\_level\_problem]\}}} \\
Triggering situation: \textcolor{high}{\texttt{\{profile[triggering\_situation]\}}} \\
Automatic thoughts: \textcolor{high}{\texttt{\{profile[automatic\_thoughts]\}}}\\

\textbf{Client behavior constraints:}\\
The client may show mild hesitation or ambivalence toward cognitive change.  
The client is aware of their automatic thoughts but may still partially endorse them.  
Cognitive change should emerge gradually, not instantly.
\\
\textbf{Counselor Instructions}\\
Warm, collaborative, and supportive.  
More directive than the understanding phase, but still gentle and respectful.

\textbf{Planning constraints}:\\
Plan for Stage~2 progress: \textcolor{high}{\texttt{\{plan\}}} \\
Action order for Stage~2: \textcolor{high}{\texttt{\{action\_order\}}}\\
\textbf{Action rules:}\\
Actions must follow the given order monotonically.  
Repeating the same action is allowed if necessary.  
No action should not be skipped.  
No actions outside the given list may be introduced.\\

\textbf{Output Format (Strict)}\\
Return the dialogue as a list of dictionaries, one dictionary per utterance.
Each dictionary must follow exactly this structure:

\texttt{\{
  "turn\_num": <int>,\\
  "role": "counselor" or "client",\\
  "action\_reasoning": "<brief reasoning; use 'n/a' for client turns>",\\
  "action": "<one action from the action order; use 'n/a' for client turns>",\\
  "utterance": "<spoken text>"\\
\}}\\

\textbf{Hard Constraints}\\
Less than  21 turns.  \\
Start with the counselor.  \\
Alternate strictly between counselor and client.  \\
End with the counselor.  \\
Use \texttt{n/a} for client action and action\_reasoning fields. \\ 
No extra commentary outside the list.

\end{prompt}

\begin{prompt}[colback=gray!5, colframe=gray!70]
{Planning Intervention Actions for Stage 2 CBT Dialogue}

You are a CBT expert therapist.
Stage~1 focuses on understanding the client, and Stage~2 focuses on performing cognitive reframing.

Your task is to take the Stage~1 dialogue history and the specified CBT strategy, and generate a structured intervention plan for the Stage~2 dialogue.

Specifically, you must generate a sequence of intervention action order keys that the counselor will follow during Stage~2.\\

\textbf{Action Constraints} \\
Each action key must satisfy the following constraints. \\ 
Each key must consist of 3--5 words.  \\
Each key must describe a specific and observable counselor action.  \\
Each key should clearly indicate what the counselor will do or ask.  \\
The final key must always be \texttt{End session}.  \\

All keys must align with the overall plan to ensure a coherent therapeutic flow.

\textbf{Input Format}

Stage~1 dialogue history:\\
\textcolor{high}{\texttt{\{history\}}}

CBT strategy (implicit in the plan generation).\\
\textcolor{high}{\texttt{\{CBT strategies\}}}\\

\textbf{Output Requirements}

The output must include the following fields.

\texttt{"plan"}:  
A short summary of the CBT strategy, explaining how the intervention plan will help the client and what therapeutic goals it aims to achieve.

\texttt{"reason\_for\_these\_order"}:  
A brief explanation of why these specific action keys were selected and why they are ordered in this sequence.

\texttt{"action\_order"}:  
A list of 5--7 action keys, where each key consists of 3--5 words and represents a concrete counselor action.\\

\textbf{Expected Output Format}

\texttt{\{\\
  "plan": "<Short description of which CBT strategy will use and therapeutic goals>",\\
  "reason\_for\_these\_order": "<Explanation of how and why the action order was designed>",\\
  "action\_order": [\\
    "restate feared weight thought",\\
    "rate belief intensity", \\
    ...
    "End session"\\
  ]\\
\}}

\end{prompt}

\begin{prompt}[colback=gray!5, colframe=gray!70]
{Filtering: Prompt for CTRS-Based Dialogue Evaluation}

You are a CBT expert trained in the Cognitive Therapy Rating Scale (CTRS).
This task uses an 8-item reduced version of CTRS.\\

Your job:
\begin{itemize}
  \item Read the session transcript carefully
  \item Assign a score from 0--6 for \textbf{each item}
  \item Base all scores strictly on the scoring definitions below
  \item Provide a JSON object with both \texttt{score} and \texttt{score\_reason} fields
  \item Do not include any text outside the JSON object
\end{itemize}

\textbf{CTRS Scoring Definitions (Use Exactly These)}

\begin{description}

\item[1. Feedback]
\begin{itemize}
  \item[0:] Therapist did not ask for feedback to determine the patient’s understanding or response.
  \item[2:] Therapist elicited some feedback but did not sufficiently check understanding or satisfaction.
  \item[4:] Therapist asked enough questions to ensure understanding and adjusted accordingly.
  \item[6:] Therapist was especially adept at eliciting and responding to feedback throughout the session.
  \item[1/3/5:] Between two adjacent descriptors.
\end{itemize}

\item[2. Understanding]
\begin{itemize}
  \item[0:] Therapist repeatedly failed to understand explicit content; poor empathy.
  \item[2:] Understood explicit content but missed subtle communication.
  \item[4:] Generally grasped the patient’s internal reality.
  \item[6:] Thoroughly understood and communicated the patient’s internal reality.
  \item[1/3/5:] Between two adjacent descriptors.
\end{itemize}

\item[3. Interpersonal Effectiveness]
\begin{itemize}
  \item[0:] Hostile, demeaning, or destructive.
  \item[2:] Interpersonal problems (impatient, aloof, insincere).
  \item[4:] Satisfactory warmth, confidence, and professionalism.
  \item[6:] Optimal interpersonal effectiveness for this patient.
  \item[1/3/5:] Between two adjacent descriptors.
\end{itemize}

\item[4. Collaboration]
\begin{itemize}
  \item[0:] No attempt at collaboration.
  \item[2:] Attempted but failed to establish rapport or shared focus.
  \item[4:] Collaborated well on an important problem.
  \item[6:] Encouraged the patient to function as an active team member.
  \item[1/3/5:] Between two adjacent descriptors.
\end{itemize}

\item[5. Guided\_discovery]
\begin{itemize}
  \item[0:] Relied on debate, persuasion, or lecturing.
  \item[2:] Overused persuasion with supportive tone.
  \item[4:] Used guided discovery appropriately.
  \item[6:] Excellent balance of questioning and intervention.
  \item[1/3/5:] Between two adjacent descriptors.
\end{itemize}

\item[6. Focusing]
\begin{itemize}
  \item[0:] Did not attempt to elicit specific cognitions or behaviors.
  \item[2:] Focused on irrelevant or unfocused areas.
  \item[4:] Focused on relevant cognitions or behaviors.
  \item[6:] Skillfully focused on key targets with high potential for progress.
  \item[1/3/5:] Between two adjacent descriptors.
\end{itemize}

\item[7. Strategy]
\begin{itemize}
  \item[0:] No CBT techniques selected.
  \item[2:] Strategy vague or unpromising.
  \item[4:] Coherent and reasonable CBT strategy.
  \item[6:] Highly promising and optimally selected CBT strategy.
  \item[1/3/5:] Between two adjacent descriptors.
\end{itemize}

\item[8. CBTtechniques (Application)]
\begin{itemize}
  \item[0:] No CBT techniques applied.
  \item[2:] CBT techniques applied with major flaws.
  \item[4:] CBT techniques applied with moderate skill.
  \item[6:] CBT techniques applied very skillfully.
  \item[1/3/5:] Between two adjacent descriptors.
\end{itemize}

\end{description}

\textbf{Session Transcript}\\
The following is the session transcript. Do \emph{not} summarize or rewrite it.\\
\textcolor{high}{\texttt{\{history\}}}\\

\textbf{Output Format (JSON only)}
\begin{verbatim}
{
  "Feedback": <0-6>,
  "Feedback_score_reason": "<reason>",
  "Understanding": <0-6>,
  "Understanding_score_reason": "<reason>",
  "Interpersonal": <0-6>,
  "Interpersonal_score_reason": "<reason>",
  "Collaboration": <0-6>,
  "Collaboration_score_reason": "<reason>",
  "Guided_discovery": <0-6>,
  "Guided_discovery_score_reason": "<reason>",
  "Focusing": <0-6>,
  "Focusing_score_reason": "<reason>",
  "Strategy": <0-6>,
  "Strategy_score_reason": "<reason>",
  "CBTtechniques": <0-6>,
  "CBTtechniques_score_reason": "<reason>"
}
\end{verbatim}

Return only this JSON object.
\end{prompt}

\begin{prompt}[colback=gray!5, colframe=gray!70]
{Filtering: Prompt for Plan–Action–Dialogue Consistency Evaluation}

You are an expert supervisor of CBT counseling dialogue systems.\\
Your task is to evaluate the \textbf{clinical quality and structural consistency}
of a counseling plan, its expanded action list, and the follow-up dialogue.\\

You must assess the materials using \textbf{three evaluation metrics} defined below.
All scores are on a \textbf{1--5 scale}.\\

\textbf{Evaluation Metrics}

\begin{description}

\item[1. Clinical\_Appropriateness]
\textbf{Definition:}\\
Evaluate how clinically appropriate and therapeutically grounded the \texttt{PLAN} is.\\
Consider:
\begin{itemize}
  \item Whether the plan correctly identifies the client’s emotional and cognitive patterns
  \item Consistency with CBT / PFA / ACT principles
  \item Whether therapeutic goals are reasonable, specific, and safe
  \item The degree to which the plan reflects understanding of the client’s needs and state
\end{itemize}

\textbf{Scoring Guide:}
\begin{itemize}
  \item[1:] Clinically inappropriate; misunderstanding of client needs or harmful direction
  \item[2:] Weak clinical grounding; vague, generic, or missing key elements
  \item[3:] Moderately appropriate; basic clinical reasoning with limited depth
  \item[4:] Strong and clinically appropriate; good grounding with minor issues
  \item[5:] Excellent; highly appropriate, well-formulated, and therapeutically robust
\end{itemize}

\item[2. Plan\_Action\_Alignment]
\textbf{Definition:}\\
Evaluate how well the \texttt{ACTION LIST} expands and operationalizes the \texttt{PLAN}.\\
Consider:
\begin{itemize}
  \item Whether actions are directly derived from the plan’s therapeutic intentions
  \item Logical expansion rather than deviation from the plan
  \item Concreteness, actionability, and clinical meaningfulness
  \item Fidelity to the plan’s core structure
\end{itemize}

\textbf{Scoring Guide:}
\begin{itemize}
  \item[1:] Poor alignment; unrelated, contradictory, or unhelpful actions
  \item[2:] Weak alignment; loosely related or poorly constructed actions
  \item[3:] Moderate alignment; general consistency with some mismatches
  \item[4:] Strong alignment; actions clearly reflect the plan with minor gaps
  \item[5:] Excellent alignment; actions precisely operationalize the plan
\end{itemize}

\item[3. Dialogue\_Adherence]
\textbf{Definition:}\\
Evaluate how well \texttt{DIAL2} adheres to the \texttt{PLAN} and \texttt{ACTION LIST}.\\
Consider:
\begin{itemize}
  \item Whether the counselor follows the intended therapeutic direction
  \item Whether actions are executed in a natural and coherent order
  \item Reflection of the plan’s priorities and stepwise structure
  \item Consistency of interventions with the defined approach
\end{itemize}

\textbf{Scoring Guide:}
\begin{itemize}
  \item[1:] No adherence; dialogue ignores or contradicts plan/actions
  \item[2:] Limited adherence; occasional alignment but mostly unfollowed
  \item[3:] Moderate adherence; partial but inconsistent implementation
  \item[4:] Strong adherence; mostly follows plan/actions with minor deviations
  \item[5:] Excellent adherence; clean and faithful implementation
\end{itemize}

\end{description}

\textbf{Input Materials}

\textbf{[Dial1: Initial dialogue used to generate plan/action]}\\
\textcolor{high}{\texttt{\{dial1\}}}\\

\textbf{[Plan]}\\
\textcolor{high}{\texttt{\{plan\}}}\\

\textbf{[Action\_List]}\\
\textcolor{high}{\texttt{\{action\}}}\\

\textbf{[Dial2: Dialogue expected to follow plan/action]}\\
\textcolor{high}{\texttt{\{dial2\}}}\\

\textbf{Output Format (JSON only)}
\begin{verbatim}
{
  "Clinical_Appropriateness": <1-5>,
  "Clinical_Appropriateness_reason": "<reason>",

  "Plan_Action_Alignment": <1-5>,
  "Plan_Action_Alignment_reason": "<reason>",

  "Dialogue_Adherence": <1-5>,
  "Dialogue_Adherence_reason": "<reason>"
}
\end{verbatim}

Return only this JSON object.
\end{prompt}

\section{Prompts for \textsc{Stepper}}
\label{app:prompt_for_stepper}
The prompts used to simulate CBT counseling.
\begin{prompt}[colback=gray!5, colframe=gray!70]
{Prompt for Simulated Client Response Generation}

You are simulating the role of a \textbf{client} in a counseling session.\\

\textbf{Client Basic Profile}\\
\textcolor{high}{\texttt{\{basic\_information\}}}\\

\textbf{Personality Traits}\\
\textcolor{high}{\texttt{\{personality\}}}\\

\textbf{Surface-Level Problem}\\
\textcolor{high}{\texttt{\{surface\_level\_problem\}}}\\

\textbf{Hidden Information (Do NOT reveal early in the session)}
\begin{itemize}
  \item Triggering situation: \textcolor{high}{\texttt{\{triggering\_situation\}}}
  \item Automatic thoughts: \textcolor{high}{\texttt{\{automatic\_thoughts\}}}
\end{itemize}

\textbf{Response Rules}
\begin{itemize}
  \item Respond \textbf{only as the client}
  \item Be natural, consistent, and emotionally authentic
  \item Do \textbf{not} reveal deeper-level information too early
  \item Do \textbf{not} step out of character
  \item Do \textbf{not} provide explanations or meta-comments
\end{itemize}

\textbf{Dialogue History}\\
\textcolor{high}{\texttt{\{dialogue\_history\}}}\\

Generate the client’s \textbf{next turn}.\\
\textcolor{high}{\texttt{\{additional\_instruction\}}}\\

\textbf{Output Format (JSON only)}
\begin{verbatim}
{
  "thoughts": "<internal thoughts>",
  "utterance": "<spoken response>"
}
\end{verbatim}

Return only this JSON object.
\end{prompt}

\begin{prompt}[colback=gray!5, colframe=gray!70]
{Prompt for Candidate Utterance Evaluation}

You are a highly skilled clinical psychologist conducting a CBT-informed counseling session.\\

\textbf{Client Profile}\\
\textcolor{high}{\texttt{\{profile\}}}\\

\textbf{Dialogue History}\\
\textcolor{high}{\texttt{\{dialogue\_history\}}}\\

\textbf{Candidate Counselor Utterances (Next Turn)}\\
The following are multiple candidate counselor utterances generated for the next turn.\\
\textcolor{high}{\texttt{\{candidates\}}}\\

\textbf{Your Task}\\
For \emph{each} candidate utterance, evaluate whether it satisfies the evaluation metric defined below.\\
Use the provided rubric to guide your judgment.\\

\textbf{Evaluation Metric and Rubric}\\
\textcolor{high}{\texttt{\{metric\_rubric\}}}\\

\textbf{Strict Output Format}\\
Return a JSON \emph{list}, where each element corresponds to exactly one candidate utterance.\\
Do \textbf{not} rewrite, modify, or paraphrase any candidate. Only evaluate them.\\

\begin{verbatim}

  {
    "metric_1": <1-5>,
    "metric_1_reason": <reason for score>,
    "metric_2": <1-5>,
    "metric_2": <reason for score>,
  ...
  }

\end{verbatim}

Return only this JSON list. Do not include any explanations or additional text.
\end{prompt}

\begin{prompt}[colback=gray!5, colframe=gray!70]
{Prompt for Candidate Plan Evaluation}

You are a highly skilled clinical psychologist specializing in CBT-based structured counseling.\\
Your task is to evaluate multiple candidate \textbf{plans} for the next therapeutic step.\\

\textbf{Dialogue History}\\
\textcolor{high}{\texttt{\{dialogue\_history\}}}\\

\textbf{Candidate Plans for the Next Step}\\
The following are multiple candidate counseling plans proposed for the next turn.\\
\textcolor{high}{\texttt{\{plan\_candidates\}}}\\

\textbf{Evaluation Metric and Rubric}\\
\textcolor{high}{\texttt{\{metric\_rubric\}}}\\

\textbf{Strict Output Format}\\
Return a JSON \emph{list}, where each entry corresponds to exactly one candidate plan.\\
Do \textbf{not} add any text outside the JSON output.\\

\begin{verbatim}
  {
    "metric_1": <1-5>,
    "metric_1_reason": <reason for score>,
    "metric_2": <1-5>,
    "metric_2": <reason for score>,
  ...
  }

\end{verbatim}

Return only this JSON list.
\end{prompt}

\section{Prompts for Evaluation}
\label{app:eval_prompt}
The prompts used for model evaluation in our experiments.
\begin{prompt}[colback=gray!5, colframe=gray!70]
{Prompt for Client-Reported Experience Evaluation}

You are an expert evaluator of psychotherapy sessions. \\

You will be provided with a transcript of a counseling session between a client and a therapist. \\
Your task is to evaluate the \textbf{client’s subjective experience after the session}, based \textbf{only} on the given conversation. \\

You must infer how the client is likely to feel at the end of the session, as if the client were completing a post-session questionnaire. \\

\textbf{Important Instructions}
\begin{itemize}
  \item Do \textbf{not} evaluate the therapist directly
  \item Do \textbf{not} summarize or describe what happened in the session
  \item Infer the client’s internal reactions and lived experience
  \item Base your judgment on the overall dialogue, not isolated turns
\end{itemize}

\textbf{Counseling Session Transcript}\\
\textcolor{high}{\texttt{\{dialogue\}}}\\

\textbf{Scoring Scale (Likert 1--5)}
\begin{itemize}
  \item 1 = Not at all
  \item 2 = Slightly
  \item 3 = Somewhat
  \item 4 = Quite a bit
  \item 5 = Very much
\end{itemize}

\textbf{Evaluation Metrics}
\begin{itemize}
  \item \textbf{Metric 1: Insight} \\
  I realized something new about myself or other people.
  
  \item \textbf{Metric 2: Perceived Support} \\
  I feel understood, supported, or reassured by my therapist.
  
  \item \textbf{Metric 3: Cognitive Distance} \\
  I feel more distanced from certain feelings, thoughts, or memories.
  
  \item \textbf{Metric 4: Empowerment} \\
  I feel more empowered, hopeful, or positive about myself.
  
  \item \textbf{Metric 5: Therapeutic Stuckness} \\
  I feel stuck, blocked, or unable to progress in therapy. \\
  \textit{(Higher score indicates greater stuckness.)}
  
  \item \textbf{Metric 6: Interpersonal Hope} \\
  I feel more positively or hopeful about another person or people.
  
  \item \textbf{Metric 7: Goal Clarity} \\
  I have become clearer about the problems or goals for me to work on.
  
  \item \textbf{Metric 8: Intervention Discomfort} \\
  I feel uncomfortable doing what my therapist is suggesting for me to do. \\
  \textit{(Higher score indicates greater discomfort.)}
  
  \item \textbf{Metric 9: Coping Skills} \\
  I feel I have improved my skills or learned new strategies to cope with my problems.
  
  \item \textbf{Metric 10: Emotional Deterioration} \\
  I now feel worse than when I started the session. \\
  \textit{(Higher score indicates worse emotional state.)}
  
  \item \textbf{Metric 11: Engagement} \\
  I feel personally invested in what I need to do in therapy to achieve my goals.
  
  \item \textbf{Metric 12: Guidance Deficit} \\
  I feel a lack of direction or guidance from my therapist. \\
  \textit{(Higher score indicates less perceived guidance.)}
  
  \item \textbf{Metric 13: Emotional Relief} \\
  I feel emotionally relieved or less burdened.
  
  \item \textbf{Metric 14: Self-Acceptance} \\
  I have accepted some aspects of myself or my situation more than before.
\end{itemize}

\textbf{Output Format (JSON only)}
\begin{verbatim}
{
  "Metric_1": {
    "score": <integer 1--5>,
    "reason": "<brief explanation grounded in the conversation>"
  },
  ...
}
\end{verbatim}

\textbf{Output Rules}
\begin{itemize}
  \item Use \textbf{all} metrics listed above
  \item Scores must be integers from 1 to 5
  \item Reasons must reference concrete cues from the dialogue
  \item Return \textbf{only} the JSON object
\end{itemize}

\end{prompt}

\begin{prompt}[colback=gray!5, colframe=gray!70]
{Prompt for Therapeutic Target Extraction}

You are given a transcript of a counseling session between a client and a therapist conducted in a Cognitive Behavioral Therapy (CBT) setting. \\

\textbf{Your task} is to extract the \textbf{main therapeutic target} discussed in the session. \\

Therapeutic targets refer to the \textbf{core cognitive or emotional elements} that the therapist and client focused on during the conversation.

\textbf{Important Instructions}
\begin{itemize}
  \item Preserve the \textbf{original wording as much as possible} when extracting the target
  \item If a target is implied but not explicitly stated, infer it \textbf{conservatively} and phrase it naturally
  \item Do \textbf{not} summarize the session or add explanatory commentary
  \item Extract \textbf{one primary therapeutic target} that best represents the session’s focus
\end{itemize}

\textbf{Counseling Session Transcript}\\
\textcolor{high}{\texttt{\{dialogue\}}}\\

\textbf{Output Format (JSON only)}
\begin{verbatim}
{
  "therapeutic_targets": "one target sentence"
}
\end{verbatim}

\textbf{Output Rules}
\begin{itemize}
  \item Return \textbf{only} the JSON object
  \item The target should be a single, concise sentence
\end{itemize}

\end{prompt}

\begin{prompt}[colback=gray!5, colframe=gray!70]
{Prompt for CBT Turn-Level Tagging (Questions and Reflections)}

You are an expert annotator trained in \textbf{CBT-informed micro-level interaction analysis}. \\

You will be provided with a \textbf{numbered, multi-turn dialogue} between a client and a counselor
(e.g., \texttt{"1 Counselor: ... \\ 2 Client: ..."}). \\

\textbf{Your task} is to analyze \textbf{ONLY the counselor’s utterances} and assign appropriate
\textbf{CBT-informed micro-action tags} based on the predefined tag sets below. \\

--------------------------------\\
\textbf{CBT QUESTION TAG SET (USE ONLY THESE)}\\
--------------------------------
\begin{itemize}
  \item \textbf{Q\_Evid}: Asking the client to identify evidence that supports or contradicts their thoughts.
  \item \textbf{Q\_Alt}: Asking the client to consider how others might interpret the same situation.
  \item \textbf{Q\_Worst}: Asking the client to describe the worst possible outcome they fear.
  \item \textbf{Q\_Util}: Asking the client to evaluate how helpful or unhelpful a thought is in real life.
  \item \textbf{Q\_Adv}: Asking the client to identify potential benefits of maintaining a thought or behavior.
  \item \textbf{Q\_Disadv}: Asking the client to identify negative consequences of holding a thought or behavior.
  \item \textbf{Q\_Real}: Asking the client to examine how well their thoughts align with observable reality.
  \item \textbf{Q\_Cont}: Asking the client to place their experience on a spectrum between two extremes.
  \item \textbf{Q\_Wish}: Asking the client to replace rigid wishes with more realistic alternatives.
  \item \textbf{Q\_Solv}: Asking the client to identify concrete problems and explore solutions.
\end{itemize}

--------------------------------\\
\textbf{CBT REFLECTION TAG SET (USE ONLY THESE)}\\
--------------------------------
\begin{itemize}
  \item \textbf{R\_Simple}: Repeating or lightly paraphrasing the client’s statement without interpretation.
  \item \textbf{R\_Emo}: Reflecting the client’s emotional or affective state.
  \item \textbf{R\_Thought}: Reflecting the client’s automatic thoughts or beliefs.
  \item \textbf{R\_Meaning}: Reflecting implied meaning or deeper significance.
  \item \textbf{R\_Reframe}: Reflecting while subtly shifting toward a more adaptive interpretation.
  \item \textbf{R\_Summary}: Synthesizing multiple client statements into a coherent reflection.
\end{itemize}

--------------------------------\\
\textbf{ANNOTATION RULES (IMPORTANT)}\\
--------------------------------
\begin{itemize}
  \item Annotate \textbf{ONLY counselor utterances}
  \item Assign tags \textbf{ONLY if the utterance functions as a question or a reflection}
  \item A single counselor utterance \textbf{may receive multiple tags}
  \item If an utterance is neither a question nor a reflection, return an empty list \texttt{[]}
  \item Base your decision on the \textbf{therapeutic function}, not surface wording
  \item Do \textbf{not} invent new tags or add explanations
\end{itemize}

--------------------------------\\
\textbf{OUTPUT FORMAT (STRICT)}\\
--------------------------------\\
Return a \textbf{Python-style dictionary} where:
\begin{itemize}
  \item Keys are counselor utterance indices: \texttt{counselor\_1}, \texttt{counselor\_2}, \dots
  \item Values are lists of tags (\texttt{Q\_*} and/or \texttt{R\_*})
\end{itemize}

\textbf{Example}
\begin{verbatim}
{
  "counselor_1": ["Q_Evid"],
  "counselor_2": ["R_Emo"],
  "counselor_3": ["Q_Alt", "Q_Real", "R_Thought"],
  "counselor_4": []
}
\end{verbatim}

--------------------------------\\
\textbf{DIALOGUE}\\
--------------------------------\\
\textcolor{high}{\texttt{\{dialogue\}}}\\

--------------------------------\\
Return \textbf{ONLY} the dictionary.

\end{prompt}

\end{document}